\documentclass[conference]{IEEEtran}
\IEEEoverridecommandlockouts
\usepackage{cite}
\usepackage{amsmath,amssymb,amsfonts}
\usepackage{algorithmic}
\usepackage{graphicx}
\usepackage{textcomp}
\usepackage{booktabs}
 \usepackage{multirow}
\usepackage{listings}
\usepackage{xcolor}
\usepackage{comment}
\usepackage{caption}
\usepackage{subfigure}
\usepackage{comment}
\definecolor{cvprblue}{rgb}{0.21,0.49,0.74}
\usepackage[T1]{fontenc}  
\usepackage{adjustbox}  
\definecolor{aliceblue}{rgb}{0.94, 0.97, 1.0}
\definecolor{champagne}{rgb}{0.97, 0.91, 0.81}
 \definecolor{cosmiclatte}{rgb}{1.0, 0.97, 0.91}
\DeclareRobustCommand{\hlsb}[1]{{\sethlcolor{cosmiclatte}\hl{#1}}}

\usepackage{cite}
\usepackage{xcolor,pifont}
\definecolor{BLUE-light}{rgb}{0.29, 0.59, 0.82}
\definecolor{BLUE-cu}{rgb}{0.7, 0.75, 0.71}
\definecolor{red-cu}{rgb}{0.9, 0.4, 0.38}
\usepackage{soul}
\usepackage{color}
\DeclareRobustCommand{\hlsent}[1]{{\sethlcolor{BLUE-cu}\hl{#1}}}
\newcommand*\colourcheck[1]{%
  \expandafter\newcommand\csname #1check\endcsname{\textcolor{#1}{\ding{52}}}%
}

\newcounter{finding}
\newcommand{\finding}[1]{\refstepcounter{finding}
 	\vspace{1mm}
	\begin{mdframed}[linecolor=gray,roundcorner=12pt,backgroundcolor=gray!15,linewidth=3pt,innerleftmargin=2pt, leftmargin=0cm,rightmargin=0cm,topline=false,bottomline=false,rightline = false]
		\textbf{Finding \arabic{finding}:} #1
	\end{mdframed}
	\vspace{3mm}
}

\usepackage{mdframed}

\colourcheck{blue}
\usepackage{hyperref}
\usepackage{xcolor}
\usepackage{colortbl}
\usepackage{amssymb}
\usepackage{pifont}
\newcommand{\xmark}{\ding{55}}%
\usepackage{xcolor}
\usepackage{soul}

\newcommand{\hlc}[2][yellow]{{%
    \colorlet{foo}{#1}%
    \sethlcolor{foo}\hl{#2}}%
}
\usepackage{bbding}
\usepackage{pifont}
\usepackage{wasysym}
\usepackage{amssymb}

\usepackage{caption}

\renewcommand\thesection{\arabic{section}}
\renewcommand\thesubsection{\thesection.\arabic{subsection}}
\renewcommand\thesubsubsection{\thesubsection.\arabic{subsubsection}}

\renewcommand\thesectiondis{\arabic{section}}
\renewcommand\thesubsectiondis{\thesectiondis.\arabic{subsection}}
\renewcommand\thesubsubsectiondis{\thesubsectiondis.\arabic{subsubsection}}

\renewcommand{\thetable}{\arabic{table}}
\renewcommand{\thefigure}{\arabic{figure}}

\hypersetup{
  colorlinks   = true, 
  urlcolor     = blue, 
  linkcolor    = red, 
  citecolor = cvprblue
}

\usepackage{xspace}
\makeatletter
\DeclareRobustCommand\onedot{\futurelet\@let@token\@onedot}
\def\@onedot{\ifx\@let@token.\else.\null\fi\xspace}

\def\eg{\emph{e.g}\onedot} 
\def\ie{\emph{i.e}\onedot} 
 
\def\etc{\emph{etc}\onedot} 
 
\def\etal{\emph{et al}\onedot}
\makeatother

\definecolor{mycommentcolor}{RGB}{225, 0, 50} 

\begin{document}

\title{How Effectively Do LLMs Extract Feature-Sentiment Pairs from App Reviews?\\
{\footnotesize }
}


\author{\IEEEauthorblockN{Faiz Ali Shah\textsuperscript{*}, Ahmed Sabir\textsuperscript{*}, Rajesh Sharma, and Dietmar Pfahl}
\IEEEauthorblockA{University of Tartu, Institute of Computer Science, Tartu, Estonia}
\thanks{\textsuperscript{*}Equal contribution.}

 }

\maketitle

\begin{abstract}
Automatic analysis of user reviews to understand user sentiments toward app functionality (\ie app features) helps align development efforts with user expectations and needs.
Recent advances in Large Language Models (LLMs) such as ChatGPT have shown impressive performance on several new tasks without updating the model's parameters \ie using zero or a few labeled examples, but the capabilities of LLMs are yet unexplored for feature-specific sentiment analysis.
The goal of our study is to explore the capabilities of LLMs to perform feature-specific sentiment analysis of user reviews. This study compares the performance of state-of-the-art LLMs, including GPT-4, ChatGPT, and different variants of Llama-2 chat, against previous approaches for extracting app features and associated sentiments in zero-shot, 1-shot, and 5-shot scenarios. 
The results indicate that GPT-4 outperforms the rule-based SAFE by 17\% in f1-score for extracting app features in the zero-shot scenario, with 5-shot further improving it by 6\%. However, the fine-tuned RE-BERT exceeds GPT-4 by 6\% in f1-score. For predicting positive and neutral sentiments, GPT-4 achieves f1-scores of 76\% and 45\% in the zero-shot setting, which improve by 7\% and 23\% in the 5-shot setting, respectively.
Our study conducts a thorough evaluation of both proprietary and open-source LLMs to provide an objective assessment of their performance in extracting feature-sentiment pairs. \footnote{\url{https://github.com/Faiz-UT/Eval-Feature-Sentiment-Extraction-LLMs}}.
\end{abstract}


\section{Introduction}
Mobile app users provide feedback on the app's functionality, usage scenarios, and desired features by submitting reviews through app marketplaces \cite{Lin2022OpinionReview,Dabrowski2022AnalysingReview}. Analyzing this feedback can help developers understand users' perceptions of app features and their evolving needs \cite{Guzman2014HowReviews}. Due to the large amount of daily feedback, manual analysis of user reviews is impractical. Several techniques have been proposed to automatically summarize user feedback to facilitate software engineering activities, such as software maintenance and evolution\cite{Dalpiaz2019RE-SWOT:Analysis,Shah2016Feature-BasedApps}.

A dedicated theme within the field of opinion mining\cite{Lin2022OpinionReview} focuses on feature-specific sentiment analysis (also called aspect-based sentiment analysis), aiming to automatically generate summaries of users' sentiments towards the functionality of software applications, commonly referred to as app features.\footnote{Our study used the terms "feature" and "app features" interchangeably}. The task of feature-specific sentiment analysis of user reviews can be decomposed into two sub-tasks. The first task, "app feature extraction", is to identify the exact words expressing the functional features\footnote{Similar to the study of \cite{Dabrowski2022AnalysingReview}, we don't consider non-functional attributes of an app as features.} of an app in a review text, and the second task, "feature-specific sentiment prediction", involves determining the sentiment (\ie positive, negative, or neutral) conveyed toward the features. For instance, in a review sentence, \textit{it is great for taking down notes}, the output of feature-specific sentiment analysis would be a feature-sentiment pair such as ("taking down notes", "positive").

Various methods have been proposed to identify app features and associated sentiments from app reviews automatically. Guzman's method \cite{Guzman2014HowReviews} involved identifying collocations as potential app features and then analyzing their associated sentiments using the SentiStrength\footnote{\url{https://github.com/zhunhung/Python-SentiStrength}} tool. Dragoni's approach\cite{dragoni2019unsupervised} uses linguistic rules based on part-of-speech patterns and Semantic Dependency Graph  (SDG) to extract app features and corresponding opinion words from user reviews. Their approach relies on lexical dictionaries to determine the sentiment polarity of opinion words. Some studies have only focused on the task of extracting app features. For this purpose, rule-based approaches like SAFE\cite{Johann2017SAFE:Reviews} or supervised methods like RE-BERT\cite{DeAraujo2021RE-BERT:Model} and T-FREX\cite{T-FREX2024} have been proposed. The results of these studies indicate that fine-tuning of pre-trained models such as BERT has significantly outperformed rule-based approaches such as SAFE for extracting app features. However, to fine-tune such models, a considerable number of task-specific training examples are still required. Furthermore, updating some of the model parameters to align with the task adds more complexity to the model fine-tuning.

\begin{figure*}[t]
  \centering
  \includegraphics[width=0.90\textwidth]{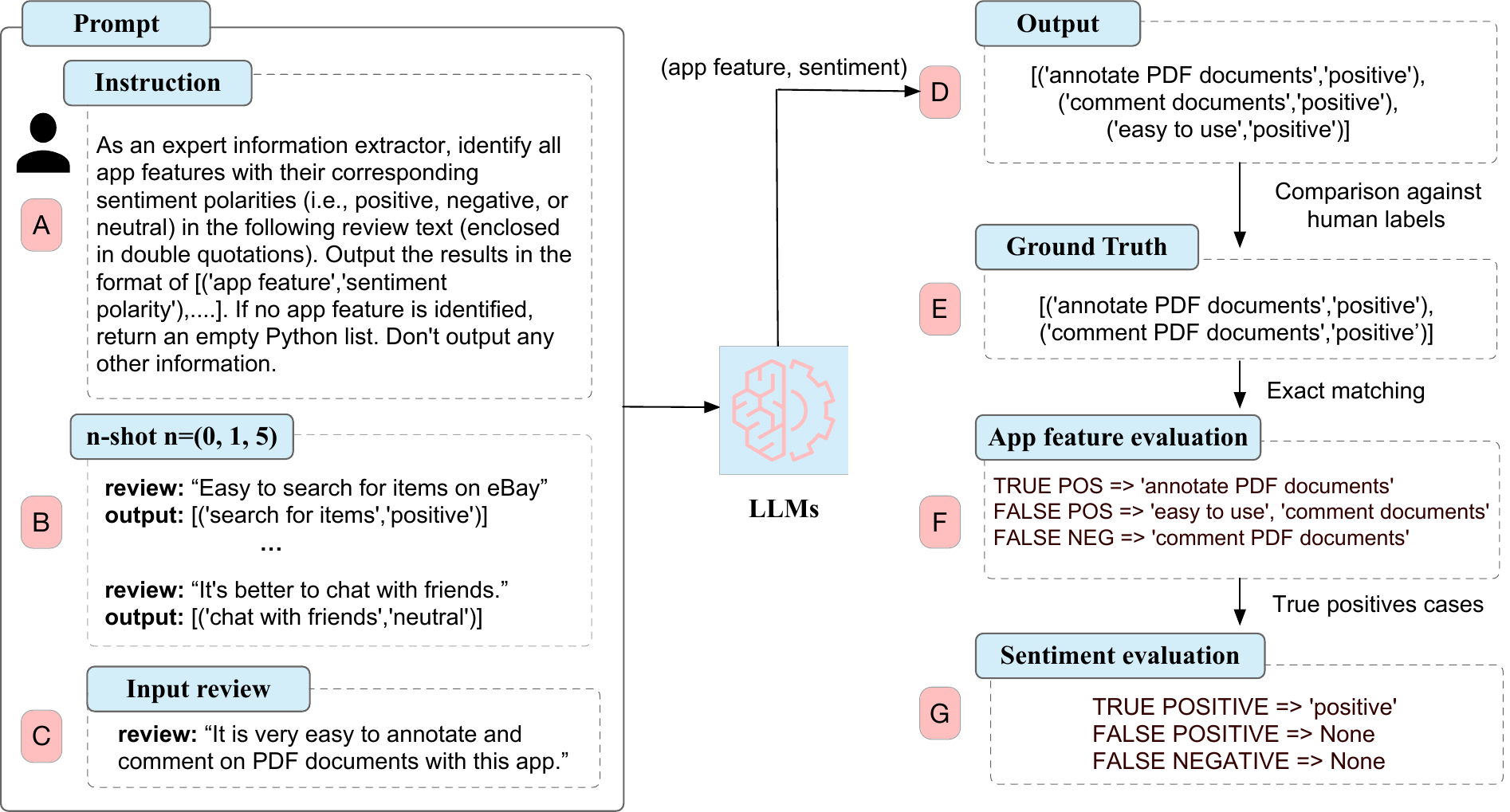}
 \vspace{-0.1cm}
\caption{Our approach for evaluating \textit{zero-shot} and \textit{few-shot} capabilities of LLMs for extracting (app feature, sentiment) pairs from a user review.}
  \label{fig:zero_few_shot_llms}
\end{figure*}

Recently, Large Language Models (LLMs) trained on a large corpus of data with Reinforcement Learning from Human Feedback (RLHF), such as ChatGPT \cite{OpenAI} and LLama-2 \cite{meta}, have shown the ability to generalize to new tasks without requiring task-specific fine-tuning\cite{liu2023summary}. LLMs take in natural language instructions defining the task (labeled as A in Figure~\ref{fig:zero_few_shot_llms}), optionally accompanied by a few examples for task demonstration (labeled as B in Figure~\ref{fig:zero_few_shot_llms}). This learning method is called \textit{zero-shot} when no examples are provided with the instruction, and \textit{few-shot} when a few examples are supplied with the task description. LLMs with RLHF have been proven effective in following instructions for different tasks with \textit{zero-shot} or \textit{few-shot} learning, eliminating the need for task-specific training examples for model parameter updating.

Zhang conducted a study\cite{Zhang_ChatGPt_Eval} that shows ChatGPT is highly effective in automatically extracting app features from app descriptions, even in \textit{zero-shot} settings. Nevertheless, app reviews can be more challenging to extract app features due to informal language, noise, and a lot of developer-irrelevant information\cite{chen2014ar}. Consequently, it remains unexplored how effectively LLMs can simultaneously extract app features and feature-specific sentiments, as demonstrated in Figure~\ref{fig:zero_few_shot_llms}. To this end, this study aims to investigate the effectiveness of LLMs for extracting features and corresponding sentiments from app reviews under \textit{zero-shot} and \textit{few-shot} settings by answering the following research questions:

\textbf{RQ1:} How does the \textit{zero-shot} performance of LLMs compare to existing methods for extracting feature-sentiment pairs from app reviews?

To address RQ1, under the \textit{zero-shot} scenario, we prompted state-of-the-art LLMs fine-tuned with RLHF, including GPT-4, ChatGPT, and various sizes of LLama-2 chat models (7B, 13B, and 70B) for extracting feature-sentiment pairs from user reviews. To evaluate the performance of LLM models, LLM predicted features and associated sentiments (labeled as 'D' in Figure~\ref{fig:zero_few_shot_llms}) are compared against human-labeled feature-sentiment pairs (labeled as 'E') using exact and partial feature matching strategies. Our results show that all LLM models have outperformed the previous rule-based approaches for extracting app features from app reviews when employing the exact match strategy. GPT-4 delivered the most impressive results, showing a substantial 23.6\% improvement in the f1-score over the best-performing rule-based approach. Regarding feature-specific sentiment prediction, GPT-4 emerged as the top-performing model, achieving an f1-score of 74\% for predicting the \textit{positive} sentiment. LLama-2-70B demonstrated the best f1-score of 50.4\% for predicting the \textit{negative} feature sentiment. For the \textit{neutral} sentiment prediction, GPT-4 outperformed other models with the best f1-score of 41\%.

\textbf{RQ2:} How does the \textit{few-shot} performance of LLMs compare to \textit{zero-shot} and existing methods for extracting feature-sentiment pairs from app reviews?

To address RQ2, we conducted an evaluation using the same LLMs as those employed in addressing RQ1, but under \textit{1-shot} and \textit{5-shot} scenarios. Our findings indicate that, compared to the \textit{zero-shot} performance, in terms of app feature extraction, the \textit{5-shot} approach led to an improvement in the f1 performance of LLama-2-70B, ChatGPT, and GPT-4 by 10.6\%, 3.5\%, and 6\%, respectively, when employing the partial match strategy with a 2-word difference. Especially, GPT-4 attained an f1-score of 56\% with the \textit{5-shot} approach, marking a significant 23\% enhancement over the SAFE approach. Furthermore, in predicting the \textit{positive} feature-specific sentiment, the \textit{5-shot} approach resulted in a 7\% increase in the f1-score for GPT-4 and a 3\% improvement for LLama-2-70B compared to their \textit{zero-shot} performances. For predicting \textit{neutral} sentiment, GPT-4 exhibited a significant gain of 23\%.

The paper is structured as follows: Section~\ref{sec:background} offers background knowledge on LLMs. Section~\ref{sec:relatd_work} summarizes related work concerning feature-specific sentiment analysis of app reviews. The details pertaining to the labeled dataset, LLM models, and their evaluation procedure are elaborated in Section~\ref{sec:experimental_settings}. The outcomes of these evaluations are presented in Section~\ref{sec:results}, followed by a discussion in Section~\ref{sec:discussion}. Section~\ref{sec:threats_to_validty} addresses the limitations and potential threats to the validity of this study. Finally, the paper concludes with a presentation of the conclusions and future avenues for research in Section~\ref{sec:conclusions}.

\section{Background}
\label{sec:background}

Large Language Models (LLMs) have experienced a remarkable surge in popularity and attention from both the general public and Natural Language Processing practitioners. These models have demonstrated significant advancements beyond the current state-of-the-art in numerous natural language tasks. These remarkable improvements led to the growing discussion of adopting such models in many everyday tasks, such as offering medical guidance, organizing job-related and a range of other applications \cite{liu2023summary}. 

LLMs have become essential in natural language processing primarily due to their ability to enhance data efficiency for targeted tasks. Fine-tuning the model parameters using gradient-based techniques for specific downstream tasks is critical to better performance. Although the fine-tuning method has yielded better performance, high-quality human-annotated label data is required.  


Another alternative method that requires no fine-tuning is In-Context Learning (ICL) \cite{brown2020language, min2022rethinking}. ICL enables LLMs to perform unseen tasks without any training by feeding a small number of examples as part of the input.


\noindent{\textbf{No Demonstration} is a \textit{zero-shot} prompting that refers to the ability of the LLMs model to perform a task without any prior knowledge related to that specific task.

\noindent{\textbf{Demonstration w/ Label} \textit{One-shot} prompting is an ICL-based demonstration with a single label that involves providing a specific example input along with a corresponding output. The task is to guide the LLMs model to generate text relevant to the given task.

\noindent{\textbf{Demonstration w/ Labels} \textit{Few-Shot} prompting is a similar approach to \textit{one-shot} ICL-based demonstration; however, the model needs to be provided with a collection of examples (\eg \textit{3-shot}, \textit{5-shot}, \etc) and a single unlabeled example, which needs to be predicted by the model.

\section{Related work}
\label{sec:relatd_work}
In this section, we discuss the previous research on feature-specific sentiment analysis of app reviews. We categorized the work into rule-based methods, supervised learning/fine-tuned language models (LMs), and Instruction-aligned human feedback (\ie RLHF) based Large Language Models (LLMs).

\subsection{Rule-based methods}
Rule-based methods use predefined patterns to extract information based on linguistic patterns or regular expressions from review text. Dragoni \etal~\cite{dragoni2019unsupervised} exploits linguistic rules consisting of part-of-speech patterns and semantic dependency relations to extract app features and their associated opinion words from a review simultaneously. To determine the sentiment polarity (\ie  positive, neutral or negative), their approach relies on lexical dictionaries. 
Guzman \etal~\cite{Guzman2014HowReviews} proposed a method identifying collocations of nouns, verbs, and adjectives as app features from app reviews, and then SentiStrength\footnote{\url{http://sentistrength.wlv.ac.uk/}} tool is applied to determine users' sentiment towards the extracted app features. 
The same approach proposed by Guzman is used by Dalpiaz \etal~\cite{Dalpiaz2019RE-SWOT:Analysis} and Shah \etal~\cite{Shah2016Feature-BasedApps} for performing feature-level sentiment analysis but for comparing competing apps. 

SUR-Miner~\cite{Gu2016WhatUsers} initially employed pre-defined templates derived from the Semantic Dependency Graph (SDG) to extract feature-opinion pairs from a review sentence. Subsequently, it utilizes the sentiment analysis tool\footnote{\url{https://nlp.stanford.edu/sentiment/}} "Deeply moving" to determine user's sentiments towards features based on opinion words.

The SAFE method is introduced by Johann \etal~\cite{Johann2017SAFE:Reviews} utilizing 18 part-of-speech and 5 sentence patterns to extract app features from user reviews. When conducting feature-specific sentiment analysis on user reviews, the SAFE method is capable only of extracting app features. Hence, an additional step is required to ascertain the sentiments of users towards the SAFE-extracted features as demonstrated in the work of \cite{Shah2019UsingSupport,Uddin2021MiningApproach,Assi2021FeatCompare:Reviews}.

In contrast to previous studies that focused on identifying specific app features and their associated sentiments, MARK framework~\cite{Vu2016MiningApproach} introduced a keyword-based approach for discovering and understanding users' opinions. Tushev's work~\cite{Tushev2022Domain-SpecificModels} presented a keyword-assisted topic modeling approach for generating coherent topics. A method called "Casper"~\cite{Guo2020Caspar:Reviews} is introduced to identify more general app issues known as "mini user stories" from review texts using parts-of-speech tagging and dependency parsing patterns. A new technique is introduced by Bakar \etal~\cite{Bakar2016ExtractingReuse} for extracting software features as phrases from user reviews to enable reusing natural language requirements. 

The rule-based approaches mentioned above are limited in their ability to identify complex patterns, domain-specific terminology, or grasp context. These limitations hinder their generalization capabilities when applied to feature-level sentiment analysis. For example,  both Dąbrowski \etal~\cite{Dabrowski2023MiningStudies} and Shah \etal~\cite{Shah2019IsStudy} discovered that the SAFE approach exhibited lower precision and recall on additional datasets compared to the results reported in the original study.

\subsection{Supervised Learning}
Supervised learning aims to train a machine learning model to learn the mapping between input features and output labels based on the labeled training data. Wang's work~\cite{Wang2022WhereUsers} treated the task of app feature extraction as Named Entity Recognition~\cite{krishnan2005named} (NER) task and built a supervised CRF model for identifying problematic features of apps from user reviews. Their results demonstrate that the supervised CRF model outperforms pattern-based feature-extraction techniques such as Caspar and SAFE. 

Fine-tuning LLMs involves taking a pre-trained model such as BERT or RoBERTa that has already been trained on a large corpus and then updating its parameters on the smaller task-specific dataset to adapt it to perform a new or more specialized task. RE-BERT~\cite{DeAraujo2021RE-BERT:Model}, a fine-tuned model, is introduced that extracts software requirements from app reviews. RE-BERT \cite{DeAraujo2021RE-BERT:Model} outperformed previous rule-based approaches such as SAFE in terms of performance with a mean score of 46\% using an exact matching strategy. A new approach named ~\cite{Wu2021IdentifyingReviews} "KEFE" is offered to identify key features from app reviews. They used SAFE rules to obtain candidate phrases and then employed a BERT-based model to classify each phrase as feature-describing or non-feature-describing phrase. Recently, a study~\cite{T-FREX2024} was conducted that fine-tuned transformer-based models such as BERT, RoBERTa, and XLNet on labeled app reviews for extracting app features, and evaluated the performance of these models under in-domain and out-of-domain settings. 

These studies suggested that fine-tuning pre-trained language models is better for identifying app features from user reviews than rule-based methods. However, such techniques require training examples for fine-tuning the model, which can be expensive to obtain. Furthermore, these models are not trained to simultaneously extract feature and sentiment pairs from user reviews.

\subsection{LLMs with RLHF}


LLMs are often used to follow instructions (prompt) to execute the user’s tasks. However, quite often, these models generate less explicit intentions than following instructions. To address this challenge, Ouyang \etal~\cite{ouyang2022training} propose InstructGPT, a method of fine-tuning a pre-trained language model through Reinforcement Learning from Human Feedback (RLHF). Specifically, the InstructGPT training starts with supervised learning based on a dataset of human-written prompts and responses. This initial phase is designed to teach the model the basic structure of understanding and generating language in a way that is aligned with human expectations. In Reinforcement Learning (RL) stage, the model receives feedback from humans to fine-tune its responses with a reward, that acts as a guide for its adjustments, ensuring they are both relevant and coherent. This iterative process enhances the model's ability to understand and meet the intricate demands of human communication.


LLMs with RLHF have demonstrated remarkable performance in a variety of downstream tasks without the need for parameter tuning. These models have shown their ability to perform various tasks by demonstrating a few examples \textit{few-shot} or by describing the task through instructions alone \textit{zero-shot} \cite{brown2020language}. Recent models (\eg ChatGPT\cite{OpenAI}, LLAMA-2 Chat \cite{meta}, \etc) build upon the foundations laid by InstructGPT, resulting in substantial improvements over LLMs trained solely on text corpora through unsupervised pre-training. 

In a recent study~\cite{Zhang_ChatGPt_Eval}, the performance of ChatGPT is evaluated to extract app features from app descriptions under \textit{zero-shot} settings. The results showed that the model achieved an 84\% precision and a 75\% recall rate on extracting app features from app descriptions. 
Building upon the encouraging results of this study~\cite{Zhang_ChatGPt_Eval}, our study aims to evaluate the performance of RLHF chat-based LLMs in extracting feature-sentiment pairs from user reviews under \textit{zero-shot} and \textit{few-shot} settings.

\section{Experimental Settings}
\label{sec:experimental_settings}

This section first describes the labeled dataset utilized to assess the performance of LLMs. Next, we provide an overview of the LLMs that have been evaluated, along with the benchmarks used for performance comparison. Finally, we describe the methodology employed to measure the performance of LLMs and elaborate on the implementation details.

\subsection{Labeled dataset}
\label{subsec:labeled_dataset}

The review dataset curated by Dąbrowski \etal~\cite{Dabrowski2023MiningStudies}, which is utilized to answer Research Questions 1 and 2 (RQ1 and RQ2), is previously used to evaluate the performances of rule-based approaches in extracting feature-sentiment pairs. Two human annotators labeled 1000 user reviews for eight different applications. The dataset comprises 1521 labeled pairs associating features with corresponding sentiments. Table~\ref{tab:stats_review_dataset} summarizes the labeled reviews and feature-sentiment statistics within this dataset. 

As we can see in Table~\ref{tab:stats_review_dataset}, the app \textit{EverNote} mentions the highest number of app features, while the app \textit{Photo Editor} mentions the lowest number of app features. 
The sentiment distribution is imbalanced in the dataset, and most features have a sentiment polarity of \textit{neutral}. The number of features with a sentiment polarity of \textit{negative} is the least.

\begin{table}[]
  \caption{Statistics of the labeled review dataset.}
   \setlength{\tabcolsep}{1.8pt} 
    \centering
  \begin{tabular}{lcccccc}
    \toprule
      & \multicolumn{2}{c}{Reviews} & \multicolumn{3}{c}{Feature-Sentiments} & Total \\
      \cmidrule(lr){2-3}\cmidrule(lr){4-6} \cmidrule(lr){7-7} 
    App & reviews & sents &  positive & neutral & negative & feature\&sentiment \\
    \midrule
    Evernote &  125 & 367 & 97 & 189 & 9 & 295 \\
    Facebook & 125 & 327  & 49 & 168 & 25 & 242 \\
    eBay & 125 & 294 &    95 & 102 & 9 & 206 \\
    Netflix & 125 & 341  & 79 & 159 & 24& 262 \\
    Spotify & 125 & 341 &  32 & 122 & 26 & 180 \\
    Photo editor & 125 & 154  & 39 & 47 & 10 & 96 \\
    Twitter & 125 & 183 & 5 & 93 & 24 & 122 \\
    WhatsApp & 125 & 169 & 20 & 84 & 14 & 118 \\
    \midrule
    \textbf{Overall} &  1000 & 2062 & 416 & 964 & 141 & 1521 \\
   \bottomrule
  \end{tabular}
  \label {tab:stats_review_dataset} 
\end{table}

\subsection{Experimented LLMs}
\label{subsec:LLMs}

We leverage the most recent state-of-the-art RLHF-based LLM models that are known for their ability to generate coherent and contextually relevant text:  




\noindent{\textbf{ChatGPT \cite{OpenAI} \& GPT-4 \cite{GPT_4}.}} follows InstructGPT \cite{ouyang2022training} training procedure via RLHF, where the model is initially adjusted using a collection of human-generated text responses. This step aims to establish a preliminary understanding and mimicry of human conversational patterns.


\noindent{\textbf{LLAMA-2 Chat \cite{meta}.} is an open-source RLHF-based LLM model that is trained on a new mix of publicly available data. Llama 2 Chat is  a LLAMA-2 version that is optimized for dialogue use cases. The model comes in different sizes (7B, 13B, and 70B).


\subsection{Baselines}
\label{subsec:baselines}
We compared the performance of our LLM-based models with four state-of-the-art baseline methods briefly described in this section. Since D\k{a}browski et al.'s study \cite{Dabrowski2023MiningStudies} evaluated these baseline methods using the same labeled dataset and evaluation procedure as our study, we used their performance results rather than re-running the baseline methods ourselves. 


\noindent{\textbf{GuMA\footnote{Same as \cite{Dabrowski2022AnalysingReview}, we refer to the approach using abbreviations derived from authors' surnames} \cite{Guzman2014HowReviews}.}} extracts app features and then determines the sentiments towards the extracted features. Both feature extraction and sentiment prediction tasks are performed independently of each other. To identify app features, GuMA uses collocations-finding algorithms that find a combination of words that co-occur frequently in the review text. For predicting sentiments, this approach first applied the SentiStrength\footnote{\url{http://sentistrength.wlv.ac.uk/}} tool that assigns a sentiment polarity score to a review sentence, and then the sentiment score is computed for the features. 

\noindent{\textbf{SAFE \cite{Johann2017SAFE:Reviews}.}} identifies app features using 18 Part-of-Speech (POS) patterns and five sentence patterns from user reviews. Since it can only extract app features in a review, an additional sentiment analysis tool needs to be applied to determine sentiment polarity towards the features. 


\noindent{\textbf{ReUS \cite{dragoni2019unsupervised}.}} uses a set of linguistic rules derived from part-of-speech patterns and semantic dependency relations to detect app features and their corresponding opinion words from a user review. Different from GuMA, REUS performed both tasks at the same time. To determine the sentiment polarity, the ReUS approach relies on lexical dictionaries. 


\noindent{\textbf{RE-BERT \cite{DeAraujo2021RE-BERT:Model}.}} posed the problem of app feature extraction from app reviews as a token classification task. This approach fine-tuned the pre-trained BERT model on the review dataset described in Section~\ref{subsec:labeled_dataset} to extract app features. RE-BERT achieved better performance for app feature extraction than the rule-based approaches GUMa, SAFE, and ReUs. Since RE-BERT doesn't extract the sentiments conveyed towards the extracted features, a separate step is required to perform feature-specific sentiment analysis. 



\subsection{Evaluation procedure}
\label{subsec:evaluation_procedure}

Since the app feature annotations can often be noisy and ambiguous, similar to previous studies\cite{Dabrowski2023MiningStudies,Shah2018TheReviews,DeAraujo2021RE-BERT:Model}, we adopted a token-based evaluation method using exact and partial matching (by difference of 1- word and 2-word) between predicted and human-annotated features. 

\noindent{\textbf{Token-based matching}} counts and evaluates each instance of an app feature separately. For instance, if an app feature \textit{voice message} occurs several times in different reviews, each feature instance will be counted separately. It might happen that depending on the context, not all \textit{voice message} bi-grams are annotated as app features. This approach distinguishes between app features and non-app features expressed with the same sequence of words. 

\noindent{\textbf{Exact matching}} requires the predicted and annotated app features to match exactly. For instance, if the annotated feature is \textit{send my message}, then to count it as a match, the predicted app feature must consist of exactly the same tokens. If the feature extraction approach extracts \textit{send message} as a feature, omitting the possessive pronoun \textit{my}, the prediction is counted as a false positive under the exact match strategy.

\noindent{\textbf{Partial matching}} allows some mismatch when comparing the predicted app features with the human-annotated features. In a partial match strategy that permits the difference of one word, the predicted feature \textit{send message} will be considered true positive even though the human-annotated app feature is \textit{send my message}. However, the predicted feature \textit{message} would be counted as a false positive because it differs from the annotated app feature in more than one word. Similarly, a partial match strategy allows the difference of two words; a predicted feature \textit{to send my message} would be counted as true positive, but a longer predicted feature such as "failed to send my message" is counted as false positive.

For evaluating the feature extraction performance of LLMs, we computed the precision, recall, and f1-score performance of LLMs (see Section~\ref{subsec:LLMs}) using exact and partial matching strategies on the whole dataset. For evaluating the sentiment prediction performance, same as the study of Dąbrowski \etal~\cite{Dabrowski2023MiningStudies}, we calculated the precision, recall, and f1-score for each sentiment polarity (\ie positive, neutral, negative) on true positive features using exact and partial matching strategies.

\subsection{Implementation Details}
\label{subsec:implementation_details}
For \textit{zero-shot} experiments, we prompted LLM models with short and long prompts without demonstrating any labeled examples. The \textit{short prompt} directly instructs the model to perform the task of feature-sentiment pair extraction from an input review.  On the other hand, the \textit{long prompt} elaborates on key concepts such as feature, feature expression, and sentiment polarity relevant to the task of feature-extraction pair, and then instructs the LLM model to proceed with the task. In this study, we denote the short prompt as \texttt{S-prompt} and the long prompt as \texttt{L-prompt}.

\begin{table*}
\caption{Comparison of \textit{zero-shot} performances of LLMs for extracting feature-sentiment pairs with \texttt{S-prompt} and \texttt{L-prompt}, and using exact and partial match strategies. (\hlc[aliceblue]{Second best result} and \hlsb{Third best result}). The (*) refers to the model fine-tuned on the dataset.}
\small
\setlength{\tabcolsep}{4pt}  
\renewcommand{\arraystretch}{1.2}  
\begin{adjustbox}{max width=\textwidth}
\begin{tabular}{lcccccccccc}
\hline 
\multirow{2}{*}{Model} & \multirow{2}{*}{Prompt type} & \multicolumn{3}{c}{Exact match $(\mathrm{n}=0)$} & \multicolumn{3}{c}{Partial match $1(n=1)$} & \multicolumn{3}{c}{Partial match $2(n=2)$}\\

  \cmidrule(lr){3-5}\cmidrule(lr){6-8}\cmidrule(lr){9-11}\
& & $\mathrm{Prec}$ & $\mathrm{Rec}$ & $\mathrm{F1}$ & $\mathrm{Prec}$ & $\mathrm{Rec}$ & $\mathrm{F1}$ & $\mathrm{Prec}$ & $\mathrm{Rec}$ & $\mathrm{F1}$ \\
\hline 
GuMa\cite{Guzman2014HowReviews}  & - & 0.05 & 0.13 & 0.07 & 0.15 & 0.37 & 0.21  & 0.18   & 0.44 & 0.25  \\
SAFE \cite{Johann2017SAFE:Reviews} & - & 0.06 & 0.06 & 0.06 & 0.24& 0.24 &  0.24 & 0.33   & 0.34 & 0.33  \\
ReUS \cite{dragoni2019unsupervised}  & - & 0.08 & 0.08 & 0.08 & 0.18 & 0.18 &  0.18 & 0.33   & 0.25 & 0.25 \\
RE-BERT* \cite{DeAraujo2021RE-BERT:Model} & - & - & - & 0.46 & - & - &  0.55 &  -  &  - &  0.62 \\
\hline 
ChatGPT \cite{OpenAI} & S & 0.227 ± 0.03 & \hlsb{0.406 ± 0.04} & \hlsb{0.290 ± 0.04} & 0.294 ± 0.05 & 0.527 ± 0.06   & \hlsb{0.377 ± 0.05} & 0.346 ± 0.04  & 0.620 ± 0.04  & \hlsb{0.443 ± 0.04}   \\
ChatGPT  & L  & 0.219 ± 0.04 & \cellcolor{aliceblue}0.433 ± 0.05 & \hlsb{0.290 ± 0.04} & 0.278 ± 0.05 & \cellcolor{aliceblue}0.551 ± 0.07   & 0.369 ± 0.06 & 0.326 ± 0.04  & \cellcolor{aliceblue}0.648 ± 0.04   &  0.433 ± 0.04 \\
 \hline 
GPT-4 \cite{GPT_4} & S & \textbf{0.257 ± 0.04} & 0.404 ± 0.06 & \cellcolor{aliceblue}0.313 ± 0.05 & \textbf{0.342 ± 0.06} & \hlsb{0.537 ± 0.07} & \textbf{0.417 ± 0.06} & \textbf{0.410 ± 0.05} & \hlsb{0.644 ± 0.06} & \textbf{0.500 ± 0.05} \\
GPT-4  & L &  \hlsb{0.240 ± 0.04}  & \textbf{0.466 ± 0.05} & \textbf{0.316 ± 0.05} & \hlsb{0.313 ± 0.07} & \textbf{0.605 ± 0.08} & \cellcolor{aliceblue}0.412 ± 0.07 & \hlsb{0.373 ± 0.06} &  \textbf{0.723 ± 0.06} & \cellcolor{aliceblue}0.491 ± 0.06 \\
 \hline
Llama-2-7B Chat \cite{meta}  & S & 0.157 ± 0.03 & 0.295 ± 0.05 & 0.205 ± 0.03 & 0.211 ± 0.03 & 0.396 ± 0.06 & 0.275 ± 0.04  & 0.255 ±  0.03 & 0.479 ± 0.05 & 0.332 ± 0.03 \\
Llama-2-7B Chat   & L  & 0.124 ± 0.01 & 0.298 ± 0.04 & 0.175 ± 0.02 & 0.168 ± 0.02 & 0.403 ± 0.05  & 0.237 ± 0.02  & 0.202 ± 0.02 & 0.485 ± 0.04 & 0.285 ± 0.02\\
Llama-2-13B Chat  & S  & 0.177 ± 0.04 & 0.265 ± 0.06  & 0.212 ± 0.05 & 0.231 ± 0.06 &0.345 ± 0.07 & 0.277 ± 0.06 & 0.280 ± 0.05  & 0.420 ± 0.07 & 0.336 ± 0.06   \\
Llama-2-13B Chat  & L  & 0.141 ± 0.02 & 0.276 ± 0.04 & 0.187 ± 0.03  & 0.190 ± 0.02 & 0.371 ± 0.05  & 0.251 ± 0.03 & 0.231 ± 0.03 & 0.452 ± 0.06 & 0.305 ± 0.03 \\
Llama-2-70B Chat  & S  & 0.218 ± 0.05 & 0.192 ± 0.03 &  0.202 ± 0.03 & 0.286 ± 0.06 & 0.254 ± 0.04  &  0.267 ± 0.04 & 0.337 ± 0.05 & 0.300 ± 0.04 & 0.314 ± 0.04  \\
Llama-2-70B Chat  & L  & \cellcolor{aliceblue}0.248 ± 0.06 & 0.273 ± 0.04 & 0.259 ± 0.05  & \cellcolor{aliceblue}0.323 ± 0.06 & 0.357 ± 0.06  & 0.338 ± 0.05 & \cellcolor{aliceblue}0.381 ± 0.05 & 0.422 ± 0.05 & 0.399 ± 0.04  \\
\hline
\end{tabular}
  \end{adjustbox}
\label{tab:zero_shot_feature_extraction_performance}
\end{table*}


To assess the performance of LLMs under \textit{few-shot} conditions, we conducted two experiments. In the first experiment, a single labeled example containing true features and sentiment labels was incorporated into the prompt (\textit{1-shot}) to guide the model in understanding the task. Subsequently, in the second experiment (\textit{5-shot}), five labeled reviews were included in the prompt to provide the models' richer contextual knowledge of the task. In both the \textit{1-shot} and \textit{5-shot} experiments, labeled examples were randomly selected from a pool of 10 review samples during each inference. These labeled examples were taken from the annotation guidelines\footnote{\url{https://github.com/jsdabrowski/IS-22}} prepared by Dąbrowski \etal~\cite{Dabrowski2023MiningStudies}.

Our prompts instruct LLM models to provide feature-sentiment pairs as a Python list of tuples or return an empty Python list if no feature is found in the input review. Nevertheless, the LLM model can produce text that either includes a Python list with additional information or excludes the Python list altogether. Hence, we employ a regular expression to extract information formatted as a Python list from the generated text of the model. In case, the regular expression couldn't find the Python list in the output, an empty Python list is returned as an output. Upon detecting a Python list within the output generated by LLMs, our script verifies its validity as a list of tuples. Each tuple within the list is expected to contain two non-empty values in string format. Furthermore, the script validates that the second value of each tuple represents a valid sentiment, such as 'positive', 'neutral', or 'negative'. Once these conditions are satisfied, the list is considered a valid collection of predicted feature-sentiment pairs.

For all \textit{zero-shot} and \textit{few-shot} experiments, we executed three iterations of each LLM model on the complete labeled dataset and calculated the average performance scores -- precision, recall, and f1-score -- alongside their standard deviation (see Section~\ref{subsec:evaluation_procedure}) for both tasks: app feature extraction and feature-specific sentiment prediction.

We utilized the OpenAI Python library\footnote{\url{https://github.com/openai/openai-python}} to prompt ChatGPT (\texttt{gpt-3.5-turbo-0613}) and GPT-4 (\texttt{gpt-4-0613})  models. For prompting LLAMA-2-Chat-models, we used models from huggingface\footnote{\url{https://huggingface.co/meta-llama}} and transformers library version 4.39.2 \footnote{\url{https://pypi.org/project/transformers/}}. Throughout all experiments,  we set the temperature $\tau$ = 0 to ensure that the model selects the highest probability in a greedy manner. Note that the temperature setting is a method to adjust the probability distribution used by the model to generate text. The \textit{max\_new\_tokens} is set to 1000.  Evaluation experiments with \texttt{Llama-2-7b-chat-hf} and \texttt{Llama-2-13b-chat-hf} were run using two NVIDIA Tesla V100 GPUs having 32 GB VRAM, while experiments with \texttt{Llama-2-70b-chat-hf} model were executed using four A100  GPUs with 80 GB VRAM. 

\section{Results}
\label{sec:results}

This section answers our RQs by comparing the \textit{zero-shot} and \textit{few-shot} performance results of LLMs with baseline methods for extracting app features and corresponding sentiments from app reviews.\\

\subsection{RQ1 - Comparison of zero-shot LLM performance and baseline methods for extracting feature-sentiment pairs}

This section presents the results of \textit{zero-shot} performance of LLMs for extracting app features, followed by the results on feature-specific sentiment prediction.

\vspace{1.18pt}

\subsubsection{\textbf{Feature extraction performance}}
Table~\ref{tab:zero_shot_feature_extraction_performance} presents the \textit{zero-shot} performance of LLM models for extracting app features from app reviews. The table shows the best average LLM performances in bold, highlighting the second and third-best LLM results. With the exact matching strategy, all LLM models outperformed rule-based approaches (\ie GuMa, SAFE, and ReUs) in all performance measures (\ie precision, recall, and f1-score). The GPT-4 has improved the average f1-score by 23.6\% over the best-performing rule-based approach ReUS. However, RE-BERT, a fine-tuned model, has shown superior performance than the best-performing GPT-4 model by 14\% in terms of average f1-score. 
 
By relaxing the matching strategy by the difference of one word, all LLM models have shown an increase in performance. The highest gain of 10\% in the average f1-score is yielded by the GPT-4 model with the \texttt{S-prompt}, and the lowest increase of 6\% in the average f1-score is shown by the LLama-2-7B model with \texttt{L-prompt}. Except for LLama-2-7B with \texttt{L-prompt}, LLM models have outperformed all rule-based approaches. Again, the highest gain of 17.7\% in the average f1-score is attained by GPT-4 with \texttt{S-prompt} over the SAFE performance (\ie  24\% f1-score).

Applying partial matching with the difference of two words has further boosted the precision and recall of all LLM models. The highest increase of 8\% in average f1-score is yielded for the GPT-4 model with \texttt{S-prompt}. GPT-4 model has shown the best average f1-score of 50\%, which is an improvement of 12\% in the average f1-score against the best-performing rule-based approach SAFE (\ie  33\%). RE-BERT (\ie  a fine-tuned model) has again shown superior performance over the best GPT-4 model by 12\% average f1-score. In our experiments, for app feature extraction, the \texttt{L-prompt} led to better recall but lower precision than the \texttt{S-prompt} across all models.

\finding{GPT-4 surpasses SAFE by 17\% in f1-score. However, the fine-tuned RE-BERT outperforms GPT-4 by 12\% in f1-score.}

\subsubsection{\textbf{Feature-specific sentiment prediction performance}}
\label{subsubsec:sentiment_rq1}
Similar to the study of Dąbrowski \etal~\cite{Dabrowski2023MiningStudies}, we only considered true positive features obtained in RQ1 and formed three datasets, each corresponding to true positive features and corresponding sentiments from the exact match, partial match $(n=1)$, and partial match $(n=2)$. 

Note that the sentiment prediction performances of LLMs are not fully comparable with ReUS and GuMa (See Section~\ref{subsec:baselines}) because LLM models have detected a higher number of true positive features. As observed in RQ1 results, loosening the match criteria increases the rate of true positive features. Therefore, in Figure~\ref{fig:zero_shot_sentiment_performance}, we compared the sentiment prediction performance of LLMs with \texttt{S-prompt} and \texttt{L-prompt} using the partial match $(n=2)$, which results in the highest number of true positive features. Figure~\ref{fig:zero_shot_sentiment_performance} shows the average sentiment prediction performances of LLMs across three runs. It is evident that performance varies depending on the sentiment polarity. The highest average f1 performance is attained for predicting \textit{positive} sentiment, followed by the \textit{negative} sentiment. The lowest average f1-score is observed for predicting \textit{neutral} sentiment. The error bars in Figure~\ref{fig:zero_shot_sentiment_performance}, representing standard deviation, highlight the greater variability in LLMs for sentiment prediction performance.

\begin{figure}[t]
  \centering
  \includegraphics[width=0.95\columnwidth]{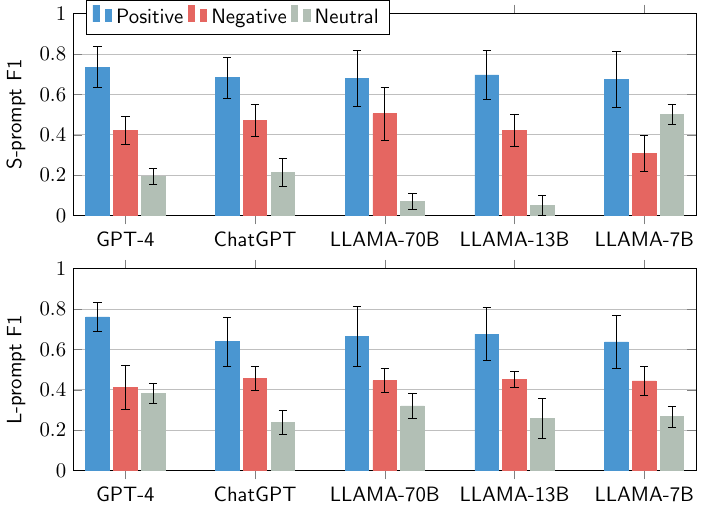}
  \vspace{-0.1cm}
  \caption{Comparison of \textit{0-shot} performances of LLMs in predicting sentiment with \texttt{S-prompt} (upper plot) and \texttt{L-prompt} (lower plot) and using partial match $(n=2)$. All models, except LLama-7B, demonstrate improved performance with the \texttt{L-prompt} for predicting \hlsent{\textit{neutral}} sentiment.}
  \label{fig:zero_shot_sentiment_performance}
\end{figure}


As we can see in Figure~\ref{fig:zero_shot_sentiment_performance}, GPT-4 model with \texttt{L-prompt} achieved the highest average f1-score of 76\% for predicting \textit{positive} sentiment.
Surprisingly, the LLama-2-70B model with the \texttt{S-prompt} has shown the best average f1-score performance for predicting negative sentiment.
The best average f1-score for the \textit{neutral} sentiment is 41\% for GPT-4 with \texttt{L-prompt}.
Interestingly, \texttt{L-prompt} improved the average f1-score performance for \textit{neutral} sentiment across all LLM models with the exception to LLama-7b. On average, the highest improvement of 19\% f1-score is yielded by LLama-2-70B, and the lowest increase of 2.4\% is shown by ChatGPT.

\finding{To predict positive and neutral sentiments, GPT-4 achieves the best f1 scores of 76.1\% and 44.8\%, respectively; while LLama-2-70B yields the best f1-score of 50.4\% for negative sentiment prediction.}


\subsection{RQ2 - Comparison of few-shot LLM performance against zero-shot and baseline methods for extracting feature-sentiment pairs}
We compare the \textit{ few-shot} (\ie \textit{1-shot} and \textit{5-shot}) performance of LLMs with zero-shot and baseline methods for extracting feature-sentiment pairs.

\begin{table*}
\caption{Comparision of \textit{zero-shot}, \textit{1-shot}, and \textit{5-shot} performances of LLMs for extracting app features from app reviews using exact and partial match strategies. (\hlc[aliceblue]{Second best result} and \hlsb{Third best result}). The (*) refers to the model fine-tuned on the dataset.}
\small
\setlength{\tabcolsep}{4pt}  
\renewcommand{\arraystretch}{1.2}  
\begin{adjustbox}{max width=\textwidth}
\begin{tabular}{lcccccccccc}
\hline 
\multirow{2}{*}{Model} & \multirow{2}{*}{Shot} & \multicolumn{3}{c}{Exact match $(\mathrm{n}=0)$} & \multicolumn{3}{c}{Partial match $1(n=1)$} & \multicolumn{3}{c}{Partial match $2(n=2)$}\\

  \cmidrule(lr){3-5}\cmidrule(lr){6-8}\cmidrule(lr){9-11}\
& & $\mathrm{Prec}$ & $\mathrm{Rec}$ & $\mathrm{F1}$ & $\mathrm{Prec}$ & $\mathrm{Rec}$ & $\mathrm{F1}$ & $\mathrm{Prec}$ & $\mathrm{Rec}$ & $\mathrm{F1}$ \\
\hline 
GuMa\cite{Guzman2014HowReviews}  & - & 0.05 & 0.13 & 0.07 & 0.15 & 0.37 & 0.21  & 0.18   & 0.44 & 0.25  \\
SAFE \cite{Johann2017SAFE:Reviews} & - & 0.06 & 0.06 & 0.06 & 0.24& 0.24 &  0.24 & 0.33   & 0.34 & 0.33  \\
ReUS \cite{dragoni2019unsupervised}  & - & 0.08 & 0.08 & 0.08 & 0.18 & 0.18 &  0.18 & 0.33   & 0.25 & 0.25 \\
RE-BERT* \cite{DeAraujo2021RE-BERT:Model} & - & - & - & 0.46 & - & - &  0.55 &  -  &  - &  0.62 \\
\hline 
 \multirow{3}{*}{ChatGPT\cite{OpenAI}} & 0 & 0.227 ± 0.03 & \hlsb{0.406 ± 0.04} & 0.290 ± 0.04 & 0.294 ± 0.05 & 0.527 ± 0.06  & 0.377 ± 0.05 & 0.346 ± 0.04  & 0.620 ± 0.04  & 0.443 ± 0.04   \\
  & 1 & 0.195 ± 0.02 & 0.402 ± 0.03 & 0.262 ± 0.03  &  0.270 ± 0.04 & \hlsb{0.556 ± 0.05}  & 0.363 ± 0.05 & 0.323 ± 0.03  &  \hlsb{0.668 ± 0.03}  & 0.434 ± 0.04 \\
  & 5 & 0.210 ± 0.03  & 0.370 ± 0.04 & 0.268 ± 0.03 & 0.305 ± 0.04 & 0.536 ± 0.04 & 0.388 ± 0.04  & 0.375 ± 0.03  & 0.662 ± 0.03 & 0.478 ± 0.03  \\
 \hline
\multirow{3}{*}{GPT-4\cite{GPT_4}}  & 0 & \hlsb{0.257 ± 0.04} & 0.404 ± 0.06 & \hlsb{0.313 ± 0.05} & \hlsb{0.342 ± 0.06} & 0.537 ± 0.07 & \hlsb{0.417 ± 0.06} & \hlsb{0.410 ± 0.05} & 0.644 ± 0.06 & \hlsb{0.500 ± 0.05} \\
& 1 & \cellcolor{aliceblue}0.272 ± 0.05 & \cellcolor{aliceblue}0.437 ± 0.07  & \cellcolor{aliceblue}0.335 ± 0.06  &  \cellcolor{aliceblue}0.354 ± 0.07 & \cellcolor{aliceblue}0.569 ± 0.09 & \cellcolor{aliceblue}0.436 ± 0.08 &  \cellcolor{aliceblue}0.417 ± 0.06 & \textbf{0.671 ± 0.06}  & \cellcolor{aliceblue}0.514 ± 0.06 \\
& 5 & \textbf{0.327 ± 0.06}  & \textbf{0.460 ± 0.06} &  \textbf{0.382 ± 0.06} & \textbf{0.414 ± 0.07} & \textbf{0.583 ± 0.07} & \textbf{0.484 ± 0.07}  & \textbf{0.480 ± 0.06} & \cellcolor{aliceblue}0.670 ± 0.05 & \textbf{0.561 ± 0.06} \\
 \hline
\multirow{3}{*}{Llama-2-7B Chat\cite{meta}} & 0 & 0.157 ± 0.03 & 0.295 ± 0.05 & 0.205 ± 0.03 & 0.211 ± 0.03 & 0.396 ± 0.06 & 0.275 ± 0.04  & 0.255 ±  0.03 & 0.479 ± 0.05 & 0.332 ± 0.03 \\
 & 1 & 0.172 ± 0.03 & 0.317 ± 0.05 & 0.223 ± 0.04 & 0.221 ± 0.03 & 0.409 ± 0.06 & 0.287 ± 0.04 & 0.269 ± 0.03 & 0.497 ± 0.05  & 0.349 ± 0.03 \\
 & 5 & 0.197 ± 0.03  & 0.334 ± 0.05 & 0.247 ± 0.04 & 0.264 ± 0.04 & 0.449 ± 0.06 & 0.332 ± 0.05 & 0.312  ± 0.03 & 0.530 ± 0.05 & 0.392  ± 0.03  \\
 \hline
\multirow{3}{*}{Llama-2-13B Chat} & 0  & 0.177 ± 0.04 & 0.265 ± 0.06  & 0.212 ± 0.05 & 0.231 ± 0.06 &0.345 ± 0.07 & 0.277 ± 0.06 & 0.280 ± 0.05  & 0.420 ± 0.07 & 0.336 ± 0.06   \\
 & 1 & 0.158 ± 0.02 & 0.310 ± 0.04 & 0.209 ± 0.03 & 0.220 ± 0.03 & 0.432 ± 0.06 & 0.291 ± 0.04 & 0.267 ± 0.02 & 0.525 ± 0.03   & 0.354 ± 0.02  \\
  & 5 & 0.186 ± 0.02  & 0.300 ± 0.03 & 0.229 ± 0.02 & 0.260 ± 0.03  & 0.419 ± 0.04 & 0.321 ± 0.03  & 0.317 ± 0.02  & 0.511 ± 0.03  & 0.391 ± 0.02 \\
\hline
\multirow{3}{*}{Llama-2-70B Chat} & 0 & 0.218 ± 0.05 & 0.192 ± 0.03 &  0.202 ± 0.03 & 0.286 ± 0.06 & 0.254 ± 0.04  &  0.267 ± 0.04 & 0.337 ± 0.05 & 0.300 ± 0.04 & 0.314 ± 0.04  \\
 & 1 & 0.171 ± 0.04 & 0.329 ± 0.07 & 0.225 ± 0.05 & 0.231 ± 0.04 & 0.443 ± 0.08 & 0.303 ± 0.05 & 0.278 ± 0.03 & 0.535 ± 0.05 & 0.366 ± 0.04  \\
 & 5 & 0.200 ± 0.03  & 0.383 ± 0.05 & 0.263 ± 0.04 & 0.268 ± 0.04  & 0.513 ± 0.07 & 0.352 ± 0.05  & 0.320 ± 0.03  & 0.614 ± 0.05 & 0.420 ± 0.03 \\
\hline

\end{tabular}
  \end{adjustbox}
\label{tab:few_shot_feature_extraction_performance}
\end{table*}

\begin{figure*}
\vspace{-0.6cm}
    \centering
    \resizebox{\textwidth}{!}{%
        \begin{tabular}{ccc}
            \subfigure[GPT-4]{\includegraphics[width=0.23\linewidth]{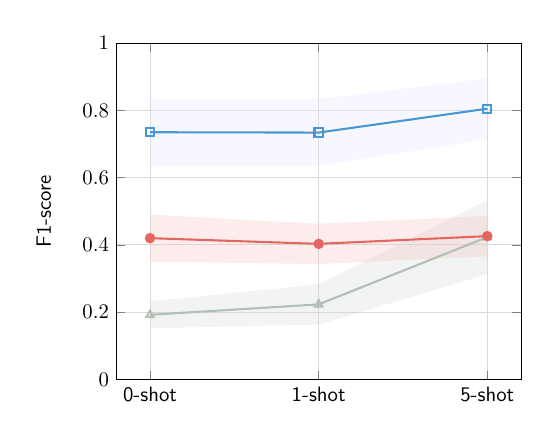}\label{fig:sub1}} &
            \subfigure[ChatGPT]{\includegraphics[width=0.21\linewidth]{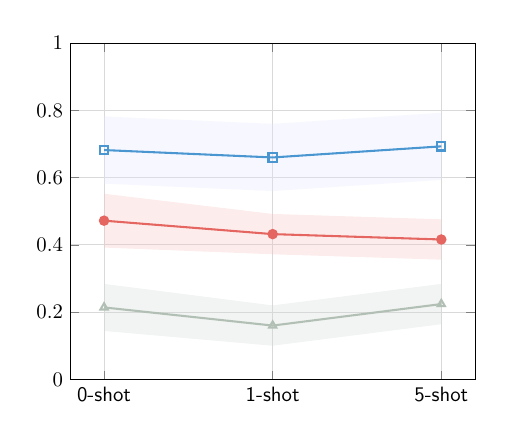}\label{fig:sub2}} &
            \subfigure[LLama-2-70B]{\includegraphics[width=0.27\linewidth]{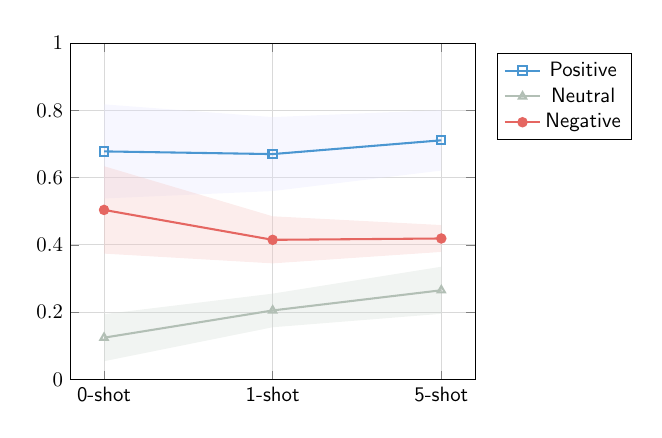}\label{fig:sub3}}
        \end{tabular}
    }
     \vspace{-0.2cm}
      \caption{Comparison of \textit{zero-shot}, \textit{1-shot}, and \textit{5-shot} performances of GPT4, ChatGPT, and LLama-70B in predicting feature-specific sentiment. \textit{5-shot} shows an increase of 7\% and 3\% in the f1-score of GPT-4 and LLama-70B for predicting \textit{positive} sentiment. For the \textit{neutral} sentiment, GPT-4 and LLama-70B f1 performance is improved by 23\% and 14\% with \textit{5-shot}.}
      \label{fig:few_shot_sentiment_performance}
    \label{fig:example}
\end{figure*}

\subsubsection{\textbf{Feature extraction performance}}
Table~\ref{tab:few_shot_feature_extraction_performance} shows the LLM performances for extracting app features from user reviews under \textit{1-shot} and \textit{5-shot} scenarios. We included the \textit{zero-shot} result from RQ1 for contextualization. Since \textit{few-shot} is to help LLMs efficiently adapt to new tasks or domains with few labeled examples, we only used \texttt{S-prompt} for our \textit{few-shot} experiments. Same as RQ1, we present the results of feature extraction with exact and partial match strategies.

For the exact match strategy, in comparison to the \textit{zero-shot} performance, \textit{1-shot} has shown a slight improvement of almost 2\% in the average f1-score of the GPT-4, LLama-7B, LLama-70B models. However, the average f1-score of the ChatGPT and LLama-13B has decreased by 2.8\% and 0.3\%, respectively. With the exception of ChatGPT, the demonstration of 5 labeled examples (\textit{5-shot}) has increased the LLM model performances; the highest gain of 6.9\% and 6.1\% is shown by GPT-4 and LLama-70B, respectively. With \textit{5-shot} performance, GPT-4 is the best LLM model with an f1-score of 38.2\%, which is 30\% better than the SAFE approach. Nevertheless, the fine-tuned RE-BERT model is 8\% better than GPT-4 in f1-score.

Compared to the \textit{zero-shot} performance, for the partial match strategy, with the difference of one word, only the average f1-score of ChatGPT is dropped by 1.4\%, while other models have shown an increase between 1\% to 3\%. The highest improvement of 3\% in the average f1-score is shown by LLama-70B. The \textit{5-shot} has boosted the performance of all LLM models, the smallest increase of 1\% f1-score is seen in ChatGPT. The highest gain of 8.5\% average f1-score over its \textit{zero-shot} performance is achieved by LLama-70B. Among all LLM models, GPT-4 is the best model with an f1-score of 48.4\%, indicating a 24\% improvement over the SAFE approach with \textit{5-shot}. Again, the fine-tuned RE-BERT model performs 7\% better than GPT-4 in terms of f1-score.

Relaxing the matching strategy to 2-word and \textit{1-shot} improved the precision and recall of LLM models, with the exception of ChatGPT. The average f1 performance of ChatGPT decreased by 1\%, but \textit{5-shot} has improved it by 3.5\%. Compared to its \textit{zero-shot} performance, with \textit{1-shot}, the LLama-70B model achieved the highest gain of 8.5\% in the f1-score, and \textit{5-shot} has further boosted it to 10.6\%. Similarly, \textit{1-shot} improved the f1 performance of LLama-7B and LLama-13B by 4\%, and \textit{5-shot} further increased it to 6\%. GPT-4 performance has increased 1.4\% in f1-score with \textit{1-shot}, while the \textit{5-shot} has increased it by 6\%. GPT-4 with \textit{5-shot} has resulted in the f1-score of 56.1\%, which is an improvement of 23\%  over the SAFE approach, but the fine-tuned RE-BERT model is still better than GPT-4 by 6\% f1-score.

\finding{With 5-shot learning, GPT-4 improved the f1-score by 6\% (i.e., 56\%), representing a 23\% improvement over the SAFE approach. However, the fine-tuned RE-BERT still outperforms GPT-4 by 6\% in f1-score.}

\subsubsection{\textbf{Feature-specific sentiment prediction performance}}
As previously described in section~\ref{subsubsec:sentiment_rq1}, we only considered the set of true positive features obtained in RQ2 and constructed three datasets, each corresponding to true positive features and corresponding sentiments from the exact match, partial match $(n=1)$, and partial match $(n=2)$. 

In line with the reasoning provided in section~\ref{subsubsec:sentiment_rq1} and constrained by limited space, Figure~\ref{fig:few_shot_sentiment_performance} compares the results of feature-specific sentiment prediction for individual sentiment polarity using the partial match $(n=2)$ only for larger models such as GPT-4, ChatGPT, and LLama-70B. 


It is evident from Figure~\ref{fig:few_shot_sentiment_performance}, for predicting the \textit{positive} sentiment, compared to \textit{zero-shot}, 
the average f1 performances of LLMs is decreased by up to 2\%. GPT-4 f1 performance almost remains the same. 
\textit{5-shot} has slightly improved the f1 performance of ChatGPT (\ie 1\%) 
but GPT-4 and LLama-70B have shown an increase of 7\% and 3\%, respectively. Under \textit{5-shot} settings, GPT-4 has achieved the best f1-score of 80.5\%  for predicting \textit{positive} sentiment. 

For the \textit{neutral} sentiment (see Figure~\ref{fig:few_shot_sentiment_performance}), ChatGPT 
performance dropped by 5\% with \textit{1-shot}, but GPT-4 and LLama-70B yielded a gain of 3\% and 8\% in average f1-score, respectively. Compared to \textit{zero-shot} performance, \textit{5-shot} has improved the performance of all LLM models. Interestingly, the average f1-score of GPT-4 and LLama-70B is increased by 23\% and 14\%, respectively, over its \textit{zero-shot} performance. Compared to \textit{zero-shot} scenario, \textit{5-shot} improved the performances 
ChatGPT 
by 1\%. Under \textit{5-shot} settings, GPT-4 has obtained the best f1-score of 42\%  for predicting \textit{neutral} sentiment.

As it can be seen in Figure~\ref{fig:few_shot_sentiment_performance}, for the \textit{negative} sentiment, implementing \textit{5-shot}, GPT-4's performance remained the same as \textit{zero-shot}, whereas \textit{1-shot} decreased its performance by 1.7\%. The f1-performance for ChatGPT decreased by 5.6\% with \textit{1-shot} and further declined by 1.6\% when \textit{5-shot} was used. With \textit{1-shot}, the highest drop of 8\% in average f1 performance is shown by LLama-70B, and with \textit{5-shot}, the f1 performance almost remains the same as the \textit{1-shot}. For the prediction of \textit{negative} sentiment, the best f1-score of almost 44\% is attained by LLama-70B with \textit{0-shot}.

\finding{For neutral sentiment prediction, 5-shot improves the f1-score by 23\% for GPT-4 and 14\% for LLama-2-70B. In positive sentiment prediction, 5-shot improves the f1-score by 7\% for GPT-4 and 3\% for LLama-70b. However, for negative sentiment, 5-shot does not improve the performance of GPT-4, and f1-score decreases for ChatGPT and LLama-2-70B by 7\% and 8\%, respectively.}



\section{Discussion}
\label{sec:discussion}



\subsection{Implications for Research}
First, this section discusses the results of  open-source and proprietary LLM models in extracting app features from user reviews. Next, we present an error analysis of LLMs using a sample of reviews. Finally, we examine the generalization capabilities of LLMs by comparing performance of LLMs against each individual app in the evaluation dataset.

\noindent{\textbf{Open-source versus Proprietary LLMs:}} Our findings show that the open-source Llama-70B model stands as the most comparable counterpart to proprietary GPT family models. 
Looking at Table~\ref{tab:few_shot_feature_extraction_performance} in Section~\ref{sec:results}, it is evident that, compared to the proprietary LLM models ChatGPT and GPT-4, the open-source LLama-2-70B shows the most substantial increase in f1-score with \textit{few-shot} learning. This underscores the potential of open-source models to approach the performance levels of closed-source ChatGPT and GPT-4 through in-context learning or knowledge distillation techniques. In our \textit{few-shot} experiments, we employed a random selection of examples for in-context learning; however, we hypothesize that selecting semantically similar examples to the input review might further enhance the performance of Llama-70B models. In addition to providing detailed instructions, our \texttt{L-prompt} employed to answer RQ1 includes examples of functional features. Interestingly, we notice that only the \textit{zero-shot} performance of LLama-70B with \texttt{L-prompt} surpasses the \textit{1-shot} performance of other LLM models by a margin of 3\% in terms of f1-score.

\begin{table*}[t!]
\centering
\caption {Feature-sentiment pairs extracted by LLama-2-70B, ChatGPT, and GPT-4 models from user reviews (\texttt{R1} to \texttt{R6}). Human-labeled features in reviews are highlighted and enclosed in brackets. \texttt{POS}, \texttt{NEU}, and \texttt{NEG} show true sentiment polarity labels in reviews.The \xmark~symbol indicates an incorrect prediction, while the \checkmark symbol indicates a correct prediction using the partial match 2 evaluation strategy.}
\resizebox{1.0\textwidth}{!}{
\begin{tabular}{llclclc}

\hline 

Review &  \multicolumn{2}{c}{LLama-2-70B Chat} & \multicolumn{2}{c}{ChatGPT} & \multicolumn{2}{c}{GPT-4}  \\ 
 \cmidrule(lr){2-3}\cmidrule(lr){4-5}\cmidrule(lr){6-7}
 & App feature &  Sentiment   & App feature & Sentiment &  App feature &  Sentiment   \\
\hline 
\texttt{R1} -> So many bugs.force crashes   & bugs \textcolor{black}{\xmark}  & \texttt{NEG} \textcolor{black}{\xmark}    & bugs \textcolor{black}{\xmark}  & \texttt{NEG} \textcolor{black}{\xmark}  &  bugs \textcolor{black}{\xmark}   &  \texttt{NEG}  \textcolor{black}{\xmark}  \\ 
     and [\hlsb{messages cannot be sent}]$_{\texttt{NEU}}$ & force crashes \textcolor{black}{\xmark}  & \texttt{NEG} \textcolor{black}{\xmark}    & force crashes \textcolor{black}{\xmark}  & \texttt{NEG} \textcolor{black}{\xmark}  &  force crashes \textcolor{black}{\xmark}   &  \texttt{NEG}  \textcolor{black}{\xmark}  \\ 
 & messages cannot be sent \textcolor{black}{\checkmark}  & \texttt{NEG} \textcolor{black}{\checkmark}    & messages cannot be sent \textcolor{black}{\checkmark}  & \texttt{NEG} \textcolor{black}{\checkmark}  &  messages cannot be sent  \textcolor{black}{\checkmark}   &  \texttt{NEG}  \textcolor{black}{\checkmark}  \\ 
\hline 
\texttt{R2} -> Best app in world & app \textcolor{black}{\xmark}  & \texttt{POS} \textcolor{black}{\xmark}    & best app \textcolor{black}{\xmark}  & \texttt{POS} \textcolor{black}{\xmark} &  None \textcolor{black}{\checkmark}   &  \texttt{None}  \textcolor{black}{\checkmark}  \\ 
\hline
\texttt{R3} -> Its easy to use and   & easy to use \textcolor{black}{\xmark}  & \texttt{POS} \textcolor{black}{\xmark}    & easy to use \textcolor{black}{\xmark}  & \texttt{POS} \textcolor{black}{\xmark}  &  easy to use \textcolor{black}{\xmark}   &  \texttt{POS}  \textcolor{black}{\xmark}  \\ 
 has a good [\hlsb{user interface}]$_{\texttt{POS}}$ & good user interface \textcolor{black}{\checkmark}  & \texttt{POS} \textcolor{black}{\checkmark}    & good user interface \textcolor{black}{\checkmark}  & \texttt{POS} \textcolor{black}{\checkmark}  &  user interface \textcolor{black}{\checkmark}   &  \texttt{POS}  \textcolor{black}{\checkmark}  \\ 
 \hline 
\texttt{R4} -> I cant [\hlsb{add filtets w/ my pics}]$_{\texttt{NEU}}$ & add filters with pictures \textcolor{black}{\xmark}  & \texttt{NEG} \textcolor{black}{\xmark}    & add filters \textcolor{black}{\xmark}  & \texttt{NEG} \textcolor{black}{\xmark}  &  add filters \textcolor{black}{\xmark}   &  \texttt{NEG}  \textcolor{black}{\xmark}  \\ 
with the lates version using my galaxy &   &     &    &   &   &   \\ 
pls fix this! &   &     &    &   &   &   \\ 
\hline 
\texttt{R5} -> Need to be [\hlsb{login}]$_{\texttt{NEU}}$ \& [\hlsb{log out}]$_{\texttt{NEU}}$  & login \& log out feature \textcolor{black}{\xmark}  & \texttt{NEU} \textcolor{black}{\checkmark}    & login \& log out feature \textcolor{black}{\xmark}  & \texttt{POS} \textcolor{black}{\xmark}  &  login \& log out feature \textcolor{black}{\xmark}   &  \texttt{POS}  \textcolor{black}{\xmark}  \\ 
feature for security reason or [\hlsb{password}]$_{\texttt{NEU}}$ & password option \textcolor{black}{\checkmark}  & \texttt{NEU} \textcolor{black}{\checkmark}    & password option \textcolor{black}{\checkmark}  & \texttt{POS} \textcolor{black}{\xmark}  &  password option \textcolor{black}{\checkmark}   &  \texttt{POS}  \textcolor{black}{\xmark}  \\ 
option &  &   &   &  & &  \\ 
\hline
\texttt{R6} -> I love it I think they should add is  & tempo/speed control \textcolor{black}{\xmark}  & \texttt{POS} \textcolor{black}{\xmark}    & add tempo/speed control \textcolor{black}{\xmark}  & \texttt{POS} \textcolor{black}{\xmark}  &  tempt/speed thing \textcolor{black}{\xmark}   &  \texttt{POS}  \textcolor{black}{\xmark}  \\ 
a tempo/speed thing so you can [\hlsb{listen at}   &   &   &  listen at different speeds \textcolor{black}{\checkmark}    & \texttt{POS} \textcolor{black}{\xmark} &  &  \\ 
\hlsb{different speeds}]$_{\texttt{NEU}}$ that would be cool. &  &   &   &  & &  \\ 
\hline 
\end{tabular}
}



\label{tbl:models_error_analysis}
\end{table*}

\noindent{\textbf{Error Analysis of LLMs:}}
As the results clearly showed, LLMs struggled in accurately extracting app features and sentiment information from user reviews. To understand the factors contributing to inaccurate feature and sentiment predictions, we randomly selected 25 reviews and examined the predictions of LLama-70B, ChatGPT, and GPT-4 models using the partial match 2 evaluation strategy. In Table~\ref{tbl:models_error_analysis}, we showcase the predictions of these models on six review sentences (\texttt{R1} to \texttt{R6}), showing the common issues encountered in feature extraction and sentiment prediction.
Within each review, features labeled by humans are highlighted and enclosed in brackets, and true sentiment polarities are indicated as \texttt{POS} (for \textit{positive}), \texttt{NEU} (for \textit{neutral}), and \texttt{NEG} (for \textit{negative}). The table shows the predictions of features or sentiments by the LLM models, with a symbol \checkmark\ indicating a correct prediction and a symbol~\xmark\ indicating an incorrect one. It is evident that LLMs encountered difficulties in understanding what constitutes an app feature, as they incorrectly identified words such as \textit{bugs}, \textit{force crashes}, and \textit{app} as features in the first and second review examples (\ie \texttt{R1} and \texttt{R2}). In the third review example \texttt{R3}, the phrase \textit{easy to use} is extracted as a potential feature, which is used to praise the non-functional aspect of the app. In the context of app feature extraction, the feature "add filtets w/ my pics" in the fourth review example \texttt{R4} was annotated by humans with the words "filtets" and "pics" (misspelled), and "w/" (an abbreviation for "with"). Interestingly, LLama-70B also extracted the same feature as human-annotated but corrected the spelling mistakes and expanded the abbreviation. Both ChatGPT and GPT-4 also extracted the same feature but corrected the spelling of "filtets" to "filters". The ability of LLMs to perform multiple tasks, such as correcting spelling and abbreviations in user reviews, requires either the creation of new evaluation benchmarks or the adoption of alternative evaluation methods distinct from traditional approaches 

In the case of review example \texttt{R4}, it seems GPT models preferred extracting shorter features such as "add filters", but in the review example \texttt{R5}, the same GPT models opted to predict longer features, such as "login \& log out feature," that span multiple human-labeled features. Looking at LLM's predictions against example user review \texttt{R6}, LLama-70B identified the feature "tempo/speed control" and ChatGPT detected the feature "add temp/speed control" but the word "control" does not exist in the review text, and both models confabulated it themselves. It is important to note that ChatGPT extracted non-consecutive feature words like "add temp/speed control" by omitting the functioning verb "is" and the article "a" from the extracted feature. Since users can express app features as non-consecutive words, this capability of LLMs is beneficial over sequence prediction models such as CRF and LSTM that can only extract consecutive words as features~\cite{Shah2018TheReviews}.

In the context of feature-specific sentiment prediction, we have observed that LLM models often confuse the sentiment \textit{neutral} with \textit{negative}. For instance, for the first review \texttt{R1} and the fourth review \texttt{R4}, labeled sentiment for features such as "messages cannot be sent" and "add filters w/ my pics" is predicted as \textit{negative}, when in reality it is \textit{neutral}. Upon investigation, it was discovered that the annotation guidelines used for review annotation instructed annotators to assign \textit{negative} or \textit{positive} sentiment only when a clear expression of \textit{positive} or \textit{negative} opinion is present towards a feature. As per these instructions, app users in the first review \texttt{R1} and fourth review \texttt{R4} stated that they could not utilize a functionality without explicitly conveying their sentiments towards these features. In the fifth example review \texttt{R5}, Llama-70B model correctly predicted the \textit{neutral} sentiment, whereas the GPT family wrongly predicted it as \textit{positive}.

\noindent{\textbf{Generalization capabilities of LLMs:}}
To assess the capabilities of LLMs in conducting feature-specific sentiment analysis across different applications, we performed a comparison of the 5-shot f1 performance scores\footnote{The f1-score for each model is calculated as the average across three feature matching strategies} of three LLM models: GPT-4, ChatGPT, and LLama-2-70B across eight different applications using a spider chart shown in Figure~\ref{fig:app_level_performance}. This analysis reveals that LLM models exhibit varying performance across different applications. Specifically, LLM models demonstrate improved performance in extracting features from user reviews of applications such as "WhatsApp," "Twitter," and "PhotoEditor". Notably, all models display a low f1-score for the "Netflix" application. This observed trend may be associated with the proportion of review data from each application employed in training or fine-tuning these models.

\subsection{Implications for Practitioners}
Our findings indicate that while LLMs show promise, their precision and recall in extracting app features from user reviews are not yet adequate for practical implementation. Although prompt engineering demonstrates lower performance compared to fine-tuned pretrained models, it offers a more cost-efficient approach for practitioners, particularly in settings where labeled data is scarce. Using LLMs for prompting is highly suited to extracting feature-sentiment data from a range of feedback sources, including in-app feedback, community forums, emails, and Reddit, which all demand a broad understanding of language.

\begin{figure}[t]
  \centering
  \includegraphics[width=0.85\columnwidth]{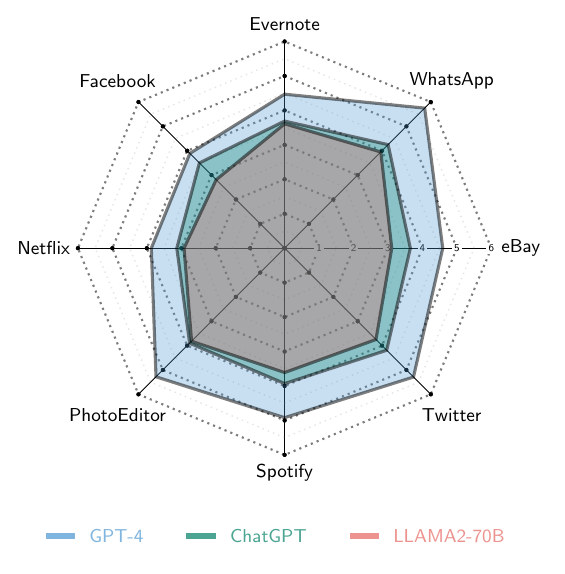}
 \vspace{-0.3cm}
  \caption{Comparison of \textit{5-shot} f1 performance of GPT-4, ChatGPT, LLama-2-70B Chat for individual app.}
  \label{fig:app_level_performance}
\end{figure}

\section{Threats to Validity}
\label{sec:threats_to_validty}

This section discusses key threats to external validity of this study by focusing on prompt engineering, the non-deterministic behavior of large language models (LLMs), the evolving nature of LLMs,  the influence of parameters on LLM performance, and the labeled review dataset.

\noindent{Prompt Engineering}: Our study used prompts to extract feature-sentiment pairs from user reviews.  We understand that the quality of the prompts used can significantly influence the results of our study. To address this potential concern, we followed the best practices of prompt design as recommended in the existing literature. Furthermore, we rigorously evaluated the effectiveness of our prompts by subjecting them to testing on sample reviews. Additionally, in addressing RQ1, we utilized two types of prompts - short and long - and performed a comparative analysis of their effectiveness in extracting app features and feature-specific sentiments. Despite presenting our findings, we acknowledge that one of the limitations of prompting LLMs is that they may struggle to understand the full context and intricacies of a given prompt, especially when it comes to highly specific or specialized information.

\noindent{Non-determinism in LLMs}: It is well known that LLMs hallucinate \cite{bang2023multitask}, and their output is non-deterministic. To mitigate this concern, we performed three iterations of each LLM on the entire review dataset to quantify the uncertainty inherent in the models' responses. We then reported the average score along with the standard deviation for each performance metric.

\noindent{LLMs Evolutionary Landscape}: Our study assessed both open-source and closed-source state-of-the-art Large Language Models (LLMs). However, the landscape of LLMs is rapidly evolving, making it impossible to evaluate all models in a single study, considering the substantial computational and financial resources needed. Therefore, it is yet to be investigated whether the results generalize to other LLMs.

\noindent{Impact of Temperature and Token Limits on LLM Performance}: In our study, all models were evaluated with the parameters temperature set to 0 and max\_new\_tokens set to 1000. As a result, further investigation is required to understand the impact of varying these parameters on the reported performance of the models for this task.

\noindent{Labeled Review Dataset}: Our evaluation study relies on a review dataset comprising 1000 labeled reviews from 8 distinct applications. We acknowledge the relatively small size of this dataset. However, to the best of our knowledge, it is the only openly available review dataset with annotated feature and sentiment pairs.

\section{Conclusions and Future Work}
\label{sec:conclusions}

Automated analysis of users' opinions regarding features in user feedback can help understand users’ perceptions of functionality. Our study has evaluated the \textit{zero-shot} and \textit{few-shot} performance of LLMs such as GPT-4, ChatGPT, and Llama-2-Chat variants, in extracting app features and associated sentiments from app reviews simultaneously. We conducted a comparative analysis of LLMs’ performances against rule-based and supervised learning methods. 

In \textit{zero-shot} learning, GPT-4 is the top-performing model for extracting app features, surpassing rule-based approaches by 23.6\% in f1-score with the exact matching strategy. However, the fine-tuned BERT model still outperforms GPT-4 by 14\% in the f1-score. In the \textit{5-shot} scenario, both GPT-4 and Llama-2-70B show further improvements of 7\% and 6\% in f1-score, respectively. 
Our evaluation study has showcased the encouraging performance of LLMs in extracting fine-grained information from user feedback. 
This capability of LLMs holds promise in assisting developers with their software maintenance and evolution activities by analyzing users' opinions.

In our future work, we aim to investigate the performance of small-scale models trained to replicate the behavior of larger, and complex models using knowledge distillation techniques \cite{gu2024minillm}. Additionally, a  Chain-of-Thought \cite{wei2022chain} (CoT) prompting strategy could enhance performance in extracting feature-sentiment information as it organizes input in a manner that mimics human reasoning. Finally, exploring parameter-efficient fine-tuning of small LLMs through the Low-Rank Adaptation \cite{hu2021lora} (\ie LoRA) technique is another promising direction for achieving human-level performance on this task.

\section*{Acknowledgment}
This work has received funding from the EU H2020 program under the SoBigData++ project (grant agreement No. 871042), CHIST-ERA grant No. CHIST-ERA-19-XAI-010, grant PRG1226 of the Estonian Research Council, and the HAMISON project.

\bibliographystyle{IEEEtran}

\appendices

\section{Detailed results of Sentiment prediction performance of LLMs with varying levels of feature-extraction matching strategies}
\label{app:appendex_a}

\begin{table*}

\caption{Comparison of sentiment prediction performances of LLMs with \texttt{S-prompt} and \texttt{L-prompt} and exact and partial feature-matching strategies.(\hlc[aliceblue]{Second best result} and \hlsb{Third best result}). Note that SAFE and RE-BERT techniques cannot predict sentiment towards app features (see Section \ref{subsec:baselines}).}
\small
\setlength{\tabcolsep}{4pt}  
\renewcommand{\arraystretch}{1.2}  
\begin{adjustbox}{max width=\textwidth}
\begin{tabular}{lccccccccccc}
\hline 

\multirow{2}{*}{Model} & \multirow{2}{*}{Prompt} &  \multirow{2}{*}{Feature matching} & \multicolumn{3}{c}{Positive sentiment} & \multicolumn{3}{c}{Neutral sentiment} & \multicolumn{3}{c}{Negative sentiment}\\

  \cmidrule(lr){4-6}\cmidrule(lr){7-9}\cmidrule(lr){10-12}\
& & & $\mathrm{Prec}$ & $\mathrm{Rec}$ & $\mathrm{F1}$ & $\mathrm{Prec}$ & $\mathrm{Rec}$ & $\mathrm{F1}$ & $\mathrm{Prec}$ & $\mathrm{Rec}$ & $\mathrm{F1}$ \\
\hline
\multirow{2}{*}{GuMa\cite{Guzman2014HowReviews}}  & - & Exact $(\mathrm{n}=0)$& 0.68 & 0.84 & 0.75 & 0.68 & 0.40 & 0.50  & 0.41   & 0.64 & 0.50  \\
  & - & Exact $(\mathrm{n}=1)$& 0.61 & 0.89 & 0.75 & 0.85 & 0.49 & 0.62  & 0.40   & 0.68 & 0.50  \\
  & - & Exact $(\mathrm{n}=2)$& 0.59  & 0.89  & 0.71 & 0.86 & 0.54 & 0.66  & 0.38   & 0.61 & 0.47  \\
\hline
\multirow{2}{*}{ReUS\cite{dragoni2019unsupervised}}  & - & Exact $(\mathrm{n}=0)$& 0.90 & 0.62 & 0.73 & 0.60 & 0.87 & 0.71  & 0.62   & 0.36 & 0.46  \\
  & - & Exact $(\mathrm{n}=1)$& 0.80 & 0.48 & 0.60 & 0.66 & 0.88 & 0.75  & 0.43   & 0.24 & 0.31  \\
  & - & Exact $(\mathrm{n}=2)$& 0.80  & 0.44 & 0.57 & 0.68 & 0.91 & 0.78  & 0.41   & 0.18 & 0.25  \\
\hline 

\multirow{2}{*}{ChatGPT\cite{OpenAI}} & \multirow{2}{*}{S} & Exact $(\mathrm{n}=0)$ & 0.515 ± 0.02  & 0.986 ± 0.02  & 0.663 ± 0.01 & 0.859 ± 0.21  & 0.133 ± 0.06  &  0.227 ± 0.10 & 0.303 ± 0.09  & 0.945 ± 0.08  & 0.452 ± 0.10   \\
 & & Partial $(\mathrm{n}=1)$ & 0.528 ± 0.11  & 0.981 ± 0.04 & 0.679 ± 0.09 & 0.956 ± 0.06 & 0.125 ± 0.04 & 0.219 ± 0.07 & 0.335 ± 0.08 & 0.958 ± 0.06  & \hlc[aliceblue]{0.492 ± 0.09} \\

 & & Partial $(\mathrm{n}=2)$  & 0.531 ± 0.12 &  0.987 ± 0.02 & 0.682 ± 0.10 & 0.950 ± 0.07 & 0.122 ± 0.04 & 0.214 ± 0.07 & 0.317 ± 0.07 & 0.957 ± 0.06 & 0.472 ± 0.08    \\
 \hline 
\multirow{2}{*}{ChatGPT} & \multirow{2}{*}{L} & Exact $(\mathrm{n}=0)$ & 0.472 ± 0.14 & \hlsb{0.994 ± 0.02} &  0.627 ± 0.11 &  0.917 ± 0.13 & 0.151 ± 0.07  &  0.252 ± 0.11  & 0.295 ± 0.08 & 0.950 ± 0.07  & 0.444 ± 0.10  \\
 & & Partial $(\mathrm{n}=1)$ & 0.481 ± 0.12 & 0.989 ± 0.02  & 0.639 ± 0.09 & 0.921 ± 0.10 & 0.140 ± 0.05 & 0.240 ± 0.09 & 0.312 ± 0.07 & 0.958 ± 0.06  & 0.466 ± 0.08 \\
 & & Partial $(\mathrm{n}=2)$ & 0.484 ± 0.13 & 0.991 ± 0.02 &  0.638 ± 0.12 & 0.915 ± 0.11 & 0.138 ± 0.04 & 0.238 ± 0.06 & 0.302 ± 0.05 & 0.951 ± 0.06 & 0.456 ±  0.06  \\
 \hline 
\multirow{2}{*}{GPT-4\cite{GPT_4}} & \multirow{2}{*}{S} & Exact $(\mathrm{n}=0)$ & 0.579 ± 0.15  &  0.973 ± 0.03 &  0.714 ± 0.12 & 0.852 ± 0.17  &  0.121 ± 0.04 & 0.210 ± 0.06  & 0.289 ± 0.10 & 0.974 ± 0.03 & 0.434 ± 0.12   \\
 & & Partial $(\mathrm{n}=1)$ &  0.596 ± 0.13 & 0.980 ± 0.03 & 0.733 ± 0.10 & 0.871 ± 0.16 & 0.110 ± 0.03 & 0.195 ± 0.05 & 0.281 ± 0.06 & 0.978 ± 0.03 & 0.432 ± 0.07\\
 & & Partial $(\mathrm{n}=2)$ & \hlsb{0.597 ± 0.12} & 0.985 ± 0.02 & \hlsb{0.735 ± 0.10} & 0.895 ± 0.12 & 0.108 ± 0.03 & 0.192 ± 0.04 & 0.270 ± 0.06 & 0.981 ± 0.02 & 0.420 ± 0.07   \\
 \hline 
\multirow{2}{*}{GPT-4} & \multirow{2}{*}{L} & Exact $(\mathrm{n}=0)$ & 0.585 ±  0.12  & 0.963 ± 0.05  & 0.719 ± 0.10 & 0.920 ± 0.08  & \textbf{0.265 ± 0.05}  & \textbf{0.408 ± 0.06} & 0.256 ± 0.07 & 0.944 ± 0.06 & 0.397 ± 0.08  \\
 & & Partial $(\mathrm{n}=1)$ & \hlc[aliceblue]{0.625 ± 0.11} & 0.972 ± 0.04 & \hlc[aliceblue]{0.754 ± 0.08} & 0.928 ± 0.07 & \hlc[aliceblue]{0.245 ± 0.05} & \hlc[aliceblue]{0.385 ± 0.06} & 0.273 ± 0.06 & 0.956 ± 0.04 & 0.421 ± 0.07\\
 & & Partial $(\mathrm{n}=2)$ & \textbf{0.629 ± 0.10} & 0.979 ± 0.02 & \textbf{0.761 ± 0.07} & 0.937 ± 0.06 & \hlsb{0.241 ± 0.04} & \hlsb{0.381 ± 0.05} & 0.266 ± 0.05 & 0.953 ± 0.05 & 0.413 ± 0.06   \\
 \hline
\multirow{2}{*}{Llama-2-7B Chat\cite{meta}} & \multirow{2}{*}{S} & Exact $(\mathrm{n}=0)$ & 0.549 ± 0.20  & 0.964 ± 0.04  & 0.679 ± 0.17  & 0.797 ± 0.15  & 0.199 ± 0.08  & 0.311 ± 0.10 &  0.295 ± 0.15 & 0.770 ± 0.23 & 0.440 ± 0.16  \\
 & & Partial $(\mathrm{n}=1)$ & 0.553 ± 0.15 & 0.976 ± 0.02 & 0.693 ± 0.13 & 0.787 ± 0.18  & 0.187 ± 0.07 & 0.297 ± 0.10 & 0.314 ± 0.10 & 0.866 ± 0.13 & 0.446 ± 0.10\\
 & & Partial $(\mathrm{n}=2)$ & 0.532 ± 0.15 & 0.974 ± 0.02 & 0.674 ± 0.14 & 0.832 ± 0.08 & 0.192 ± 0.06  & 0.306 ± 0.09 & 0.309 ± 0.07 & 0.850 ± 0.12 & 0.444 ± 0.07  \\
\hline
\multirow{2}{*}{Llama-2-7B Chat} & \multirow{2}{*}{L} & Exact $(\mathrm{n}=0)$ & 0.481 ± 0.16  & 0.980 ±  0.03 & 0.627 ±  0.16 & 0.858 ±  0.13  & 0.161 ±  0.06  & 0.263 ±  0.08 & 0.323 ±  0.10  & 0.781 ±  0.25 & 0.415 ±  0.10   \\
 & & Partial $(\mathrm{n}=1)$ & 0.490 ±  0.13  & 0.984 ±  0.02 & 0.643 ±  0.13 & 0.860 ±  0.10  & 0.151 ±  0.06  & 0.250 ±  0.09 & 0.310 ±  0.11 & 0.798 ±  0.21 & 0.408 ± 0.08\\
 & & Partial $(\mathrm{n}=2)$ &  0.485 ±  0.13 & 0.980 ±  0.02 & 0.637 ±  0.12 & 0.870 ±  0.09  & 0.160 ±  0.03 & 0.266 ±  0.05  & 0.306 ±  0.11 & 0.801 ±  0.20 & 0.413 ±  0.11   \\
  \hline
\multirow{2}{*}{Llama-2-13B Chat} & \multirow{2}{*}{S} & Exact $(\mathrm{n}=0)$ & 0.532 ± 0.17  & 0.968 ± 0.03  & 0.668 ± 0.15  & 0.432 ± 0.33  & 0.043 ± 0.05 & 0.077 ± 0.09 & 0.272 ± 0.12  & \hlc[aliceblue]{0.994 ± 0.02}  & 0.413 ± 0.14  \\
 & & Partial $(\mathrm{n}=1)$ & 0.562 ± 0.14 & 0.975 ± 0.02 & 0.701 ± 0.13 & 0.497 ± 0.27  & 0.047 ± 0.04 & 0.085 ± 0.07 & 0.279 ± 0.09 & \textbf{0.994 ± 0.01} & 0.428 ± 0.10   \\
 & & Partial $(\mathrm{n}=2)$ & 0.552 ± 0.13 & 0.980 ± 0.01 & 0.696 ± 0.12 & 0.598 ± 0.28 & 0.051 ± 0.05 & 0.092 ± 0.08 & 0.273 ± 0.06 & \hlsb{0.990  ± 0.02} & 0.423 ± 0.08    \\
\hline
\multirow{2}{*}{Llama-2-13B Chat} & \multirow{2}{*}{L} & Exact $(\mathrm{n}=0)$ & 0.496 ± 0.16  & 0.973 ± 0.03  & 0.640 ± 0.15  &  0.870 ± 0.11 & 0.163 ± 0.09  & 0.263 ± 0.12  & \hlsb{0.340 ± 0.10}  & 0.947 ± 0.05 & 0.487 ± 0.12  \\
 & & Partial $(\mathrm{n}=1)$ & 0.529 ± 0.15 & 0.976 ± 0.03 & 0.672 ± 0.13 & 0.883 ± 0.10  & 0.148 ± 0.08 & 0.243 ± 0.11 & 0.286 ± 0.05 & 0.950 ± 0.05 & 0.436 ± 0.06\\
 & & Partial $(\mathrm{n}=2)$ & 0.533 ± 0.15 & 0.981 ± 0.02 & 0.676 ± 0.13 & 0.897 ± 0.09  & 0.156 ± 0.07 & 0.257 ± 0.10 & 0.297 ± 0.04 & 0.949 ± 0.05 & 0.450 ± 0.04  \\
\hline
\multirow{2}{*}{Llama-2-70B Chat} & \multirow{2}{*}{S} & Exact $(\mathrm{n}=0)$ & 0.513 ± 0.21  & 0.955 ± 0.11  & 0.647 ± 0.21  & 0.722 ± 0.27  & 0.074 ± 0.04  & 0.131 ± 0.07 & 0.313 ± 0.22  & 0.880 ± 0.22 & 0.428 ± 0.23  \\
 & & Partial $(\mathrm{n}=1)$ & 0.541 ± 0.17 & \hlc[aliceblue]{0.997 ± 0.01} & 0.684 ± 0.15 & 0.722 ± 0.27  & 0.062 ± 0.04 & 0.113 ± 0.07 & \hlc[aliceblue]{0.341 ± 0.14} & 0.968 ± 0.07 & \hlsb{0.485 ± 0.15}\\
 & & Partial $(\mathrm{n}=2)$ & 0.530 ± 0.16 & \textbf{0.998 ± 0.01} &  0.678 ± 0.14 & 0.727 ± 0.28 & 0.069 ± 0.04 & 0.124 ± 0.07 & \textbf{0.355 ± 0.13} & 0.967 ± 0.08 & \textbf{0.504 ± 0.13}  \\
\hline
\multirow{2}{*}{Llama-2-70B Chat} & \multirow{2}{*}{L} & Exact $(\mathrm{n}=0)$ & 0.507 ± 0.17  & 0.992 ± 0.02  & 0.652 ± 0.16  & \textbf{0.984 ± 0.02}  & 0.199 ± 0.07  & 0.323 ± 0.09 & 0.326 ± 0.15 & 0.969 ± 0.04  & 0.465 ± 0.15   \\
 & & Partial $(\mathrm{n}=1)$ & 0.522 ± 0.17 & 0.983 ± 0.04 & 0.662 ± 0.16 & \hlsb{0.972 ± 0.04}  & 0.197 ± 0.04 & 0.325 ± 0.05 & 0.292 ± 0.07 & 0.978 ± 0.03 & 0.444 ± 0.08\\
 & & Partial $(\mathrm{n}=2)$ & 0.519 ± 0.15 & 0.989 ± 0.02 & 0.664 ± 0.15 & \hlc[aliceblue]{0.976 ± 0.03}  & 0.194 ± 0.04 & 0.320 ± 0.06 & 0.292 ± 0.06 & 0.980 ± 0.02 & 0.446 ± 0.06  \\
\hline
\end{tabular}
\label{tbl:zero_shot_sentiment_prediction}
  \end{adjustbox}
\end{table*}

\begin{table*}[ht]
\centering
\small 
\begin{tabular}{@{}ll@{}}
\toprule
 Prompt  & Type     \\ \midrule %
As an expert information extractor, identify all app features  &  Short \\ 
with their corresponding  sentiment polarities (i.e., positive, negative or neutral)  &  \\ 
in the following review text (enclosed in double quotations). &   \\ 
 Output the results in the format of  [('app feature', 'sentiment polarity'), ...]. &   \\ 
If no app feature is identified, return an empty Python list. & \\ 
Don't output any other information. & \\ 

\\
\hline
\\
Consider the following definitions of "feature", "feature expression",   &  Long \\ 

and "sentiment polarity": The "feature" refers to a software application functionality  &  \\ 

(e.g., "send message"), a module (e.g., "user account") providing functionalities &  \\ 

(e.g., "delete account" or "edit information") or a design component & \\ 

(e.g., UI) providing functional capabilities (e.g., "configuration screen", "button"). & \\ 

The "feature expression" is an actual sequence of words that appears in a review & \\ 
text and explicitly indicate a feature. The "sentiment polarity" refers to the & \\ 

degree of positivity, negativity or neutrality expressed towards the feature of a software & \\ 
application, and the available polarities includes: 'positive', 'neutral' or 'negative'. & \\ 

As an expert information extractor, identify all feature expressions with their corresponding \\  
sentiment polarities in  the following review text (enclosed in double quotations). Output the  & \\  

results in the format of  [('feature expression','sentiment polarity'), ...]. If no feature expression & \\ 
  is identified, return the empty Python list. Don't output any other information. & \\ 

\\ 

\addlinespace
\hline  \addlinespace
\end{tabular}
\vspace{-0.3cm}
\caption{Prompt type}
\label{prompts}
\end{table*}

\begin{table*}
\label{comparision_sentiment_prediction_rq2}
\caption{Comparison of \textit{zero-shot}, \textit{1-shot}, \textit{5-shot} performances of LLMs for predicting feature-specific sentiments using exact and partial feature matching strategies. (\hlc[aliceblue]{Second best result} and \hlsb{Third best result}). Note that SAFE and RE-BERT techniques cannot predict sentiment towards app features (see Section \ref{subsec:baselines}).}
\small
\setlength{\tabcolsep}{4pt}  
\renewcommand{\arraystretch}{1.2}  
\begin{adjustbox}{max width=\textwidth}
\begin{tabular}{lccccccccccc}
\hline 
\multirow{2}{*}{Model} & \multirow{2}{*}{Shot} &  \multirow{2}{*}{Feature matching} & \multicolumn{3}{c}{Positive sentiment} & \multicolumn{3}{c}{Neutral sentiment} & \multicolumn{3}{c}{Negative sentiment}\\

  \cmidrule(lr){4-6}\cmidrule(lr){7-9}\cmidrule(lr){10-12}\
& & & $\mathrm{Prec}$ & $\mathrm{Rec}$ & $\mathrm{F1}$ & $\mathrm{Prec}$ & $\mathrm{Rec}$ & $\mathrm{F1}$ & $\mathrm{Prec}$ & $\mathrm{Rec}$ & $\mathrm{F1}$ \\
\hline
\multirow{2}{*}{GuMa\cite{Guzman2014HowReviews}}  & - & Exact $(\mathrm{n}=0)$& 0.68 & 0.84 & 0.75 & 0.68 & 0.40 & 0.50  & 0.41   & 0.64 & 0.50  \\
  & - & Exact $(\mathrm{n}=1)$& 0.61 & 0.89 & 0.75 & 0.85 & 0.49 & 0.62  & 0.40   & 0.68 & 0.50  \\
  & - & Exact $(\mathrm{n}=2)$& 0.59  & 0.89  & 0.71 & 0.86 & 0.54 & 0.66  & 0.38   & 0.61 & 0.47  \\
\hline
\multirow{2}{*}{ReUS\cite{dragoni2019unsupervised}}  & - & Exact $(\mathrm{n}=0)$& 0.90 & 0.62 & 0.73 & 0.60 & 0.87 & 0.71  & 0.62   & 0.36 & 0.46  \\
  & - & Exact $(\mathrm{n}=1)$& 0.80 & 0.48 & 0.60 & 0.66 & 0.88 & 0.75  & 0.43   & 0.24 & 0.31  \\
  & - & Exact $(\mathrm{n}=2)$& 0.80  & 0.44 & 0.57 & 0.68 & 0.91 & 0.78  & 0.41   & 0.18 & 0.25  \\
\hline 
\hline 
\multirow{2}{*}{ChatGPT\cite{OpenAI}} & \multirow{2}{*}{0} & Exact$(\mathrm{n}=0)$ & 0.515 ± 0.02  & 0.986 ± 0.02  & 0.663 ± 0.01 & 0.859 ± 0.21  & 0.133 ± 0.06  &  0.227 ± 0.10 & 0.303 ± 0.09  & 0.945 ± 0.08  & 0.452 ± 0.10   \\
 & & Partial $(\mathrm{n}=1)$ & 0.528 ± 0.11  & 0.981 ± 0.04 & 0.679 ± 0.09 & 0.956 ± 0.06 & 0.125 ± 0.04 & 0.219 ± 0.07 & \hlsb{0.335 ± 0.08} & 0.958 ± 0.06  & \hlc[aliceblue]{0.492 ± 0.09} \\
 & & Partial $(\mathrm{n}=2)$  & 0.531 ± 0.12 &  0.987 ± 0.02 & 0.682 ± 0.10 & 0.950 ± 0.07 & 0.122 ± 0.04 & 0.214 ± 0.07 & 0.317 ± 0.07 & 0.957 ± 0.06 & 0.472 ± 0.08    \\
 \hline
\multirow{2}{*}{ChatGPT} & \multirow{2}{*}{1} & Exact $(\mathrm{n}=0)$ &  0.485 ±  0.15  & \hlsb{0.999 ±  0.00}  & 0.637 ±  0.14  & 0.925 ±  0.11  & 0.108 ±  0.05  & 0.189 ±  0.08 &  0.249 ±  0.04 & 0.948 ±  0.08 & 0.391 ±  0.06  \\
 & & Partial $(\mathrm{n}=1)$ & 0.494 ±  0.12 & \hlc[aliceblue]{0.999 ±  0.00} & 0.651 ±  0.10 & 0.930 ±  0.10  & 0.094 ±  0.03 & 0.168 ± 0.06 & 0.276 ±  0.03 & 0.959 ±  0.06 & 0.427 ±  0.04 \\
 & & Partial $(\mathrm{n}=2)$ & 0.501 ±  0.11 & \textbf{0.999 ±  0.00} & 0.660 ±  0.10 & 0.935 ±  0.09 & 0.089 ± 0.03  & 0.160 ±  0.06 & 0.280 ±  0.05 & 0.961 ±  0.06 & 0.432 ±  0.06  \\
\hline
\multirow{2}{*}{ChatGPT} & \multirow{2}{*}{5} & Exact $(\mathrm{n}=0)$ &  0.511 ±  0.15 & 0.996 ±  0.01  & 0.660 ±  0.13  & 0.912 ±  0.12  & 0.144 ±  0.05  & 0.245 ±  0.08 & 0.236 ±  0.06  & 0.944 ±  0.07 & 0.370 ±  0.08  \\
 & & Partial $(\mathrm{n}=1)$ & 0.537 ±  0.13 & 0.997 ±  0.01 & 0.690 ±  0.11 & 0.940 ±  0.08  & 0.136 ±  0.04 & 0.235 ±  0.07 & 0.268 ±  0.05 & 0.963 ±  0.05 & 0.416 ±  0.06 \\
 & & Partial $(\mathrm{n}=2)$ & 0.540 ±  0.12 & 0.998 ±  0.01 & 0.693 ±  0.10 & 0.946 ±  0.07 & 0.128 ±  0.03 & 0.224 ±  0.06 & 0.268 ±  0.05 & 0.968 ±  0.04 & 0.416 ±  0.06  \\
\hline
\hline
\multirow{2}{*}{GPT-4\cite{GPT_4}} & \multirow{2}{*}{0} & Exact $(\mathrm{n}=0)$ & 0.579 ± 0.15  &  0.973 ± 0.03 &  0.714 ± 0.12 & 0.852 ± 0.17  &  0.121 ± 0.04 & 0.210 ± 0.06  & 0.289 ± 0.10 & 0.974 ± 0.03 & 0.434 ± 0.12   \\
 & & Partial $(\mathrm{n}=1)$ &  0.596 ± 0.13 & 0.980 ± 0.03 & 0.733 ± 0.10 & 0.871 ± 0.16 & 0.110 ± 0.03 & 0.195 ± 0.05 & 0.281 ± 0.06 & 0.978 ± 0.03 & 0.432 ± 0.07\\
 & & Partial $(\mathrm{n}=2)$ & 0.597 ± 0.12 & 0.985 ± 0.02 & \hlc[aliceblue]{0.735 ± 0.10} & 0.895 ± 0.12 & 0.108 ± 0.03 & 0.192 ± 0.04 & 0.270 ± 0.06 & 0.981 ± 0.02 & 0.420 ± 0.07   \\
 \hline
\multirow{2}{*}{GPT-4} & \multirow{2}{*}{1} & Exact $(\mathrm{n}=0)$ & 0.595 ±  0.19  & 0.981 ±  0.03  & 0.719 ±  0.17  & 0.946 ±  0.19  & 0.153 ±  0.06  & 0.260 ±  0.08 & 0.272 ±  0.09  & 0.985 ±  0.03 & 0.419 ±  0.10  \\
 & & Partial $(\mathrm{n}=1)$ & 0.599 ± 0.15  & 0.986 ±  0.02 & 0.732 ±  0.13 & 0.940 ±  0.10  & 0.136 ±  0.05 & 0.235 ±  0.07 & 0.268 ±  0.05 & 0.986 ±  0.02 & 0.419 ±  0.06\\
 & & Partial $(\mathrm{n}=2)$ & 0.596 ± 0.13 & 0.990 ±  0.02 & \hlsb{0.734 ± 0.11} & 0.948 ±  0.09 & 0.128 ±  0.04 & 0.223 ±  0.06 & 0.255 ± 0.05  & 0.983 ±  0.03 & 0.403 ±  0.06  \\
\hline
\multirow{2}{*}{GPT-4} & \multirow{2}{*}{5} & Exact $(\mathrm{n}=0)$ &  \hlsb{0.686 ±  0.14}  &  0.976 ±  0.03  & 0.797 ±  0.10  & \textbf{0.963 ±  0.05}  & \textbf{0.298 ±  0.11}  & \textbf{0.444 ±  0.12} & 0.295 ±  0.08  & 0.982 ±  0.03 & 0.446 ±  0.09  \\
 & & Partial $(\mathrm{n}=1)$ & \hlc[aliceblue]{0.688 ± 0.13} &  0.979 ±  0.02 & 0.800 ±  0.10 & \hlsb{0.959 ± 0.04} & \hlsb{0.275 ± 0.10}  & \hlsb{0.416 ±  0.12} & 0.288 ± 0.05 & 0.986 ±  0.02 & 0.443 ±  0.06  \\
 & & Partial $(\mathrm{n}=2)$ & \textbf{0.691 ±  0.12} & 0.984 ±  0.02 & \textbf{0.805 ±  0.09} & \hlc[aliceblue]{0.960 ± 0.04} & \hlc[aliceblue]{0.279 ±  0.10}  & \hlc[aliceblue]{0.423 ± 0.11} & 0.274 ±  0.04 & 0.981 ±  0.03 & 0.426 ±  0.06  \\
\hline
 \hline
\multirow{2}{*}{Llama-2-7B Chat\cite{meta}} & \multirow{2}{*}{0} & Exact $(\mathrm{n}=0)$ & 0.549 ± 0.20  & 0.964 ± 0.04  & 0.679 ± 0.17  & 0.797 ± 0.15  & 0.199 ± 0.08  & 0.311 ± 0.10 &  0.295 ± 0.15 & 0.770 ± 0.23 & 0.440 ± 0.16  \\
 & & Partial $(\mathrm{n}=1)$ & 0.553 ± 0.15 & 0.976 ± 0.02 & 0.693 ± 0.13 & 0.787 ± 0.18  & 0.187 ± 0.07 & 0.297 ± 0.10 & 0.314 ± 0.10 & 0.866 ± 0.13 & 0.446 ± 0.10\\
 & & Partial $(\mathrm{n}=2)$ & 0.532 ± 0.15 & 0.974 ± 0.02 & 0.674 ± 0.14 & 0.832 ± 0.08 & 0.192 ± 0.06  & 0.306 ± 0.09 & 0.309 ± 0.07 & 0.850 ± 0.12 & 0.444 ± 0.07  \\
 \hline
\multirow{2}{*}{Llama-2-7B Chat} & \multirow{2}{*}{1} & Exact $(\mathrm{n}=0)$ & 0.531 ±  0.16  & 0.942 ±  0.05  & 0.665 ±  0.14  & 0.761 ±  0.12  & 0.177 ±  0.07  & 0.281 ±  0.09 & 0.321 ±  0.12  & 0.820 ±  0.15 & 0.434 ±  0.10  \\
 & & Partial $(\mathrm{n}=1)$ & 0.552 ±  0.12 & 0.957 ±  0.03 & 0.691 ±  0.10 & 0.775 ±  0.13  & 0.175 ±  0.07 & 0.280 ±  0.10 & 0.314 ±  0.07 & 0.869 ±  0.14 & 0.446 ±  0.05\\
 & & Partial $(\mathrm{n}=2)$ & 0.539 ±  0.11 & 0.962 ±  0.03 & 0.684 ±  0.09 & 0.781 ±  0.11 & 0.170 ±  0.06 & 0.274 ±  0.09 & 0.304 ±  0.06 & 0.863 ±  0.12 & 0.439 ±  0.06  \\
 \hline
\multirow{2}{*}{Llama-2-7B Chat} & \multirow{2}{*}{5} & Exact $(\mathrm{n}=0)$ &  0.538 ± 0.17  & 0.925 ± 0.05  & 0.659 ± 0.15  & 0.736 ± 0.19  & 0.208 ± 0.06  &  0.321 ± 0.09 & 0.272 ± 0.11  & 0.804 ± 0.17  & 0.389 ± 0.12  \\
 & & Partial $(\mathrm{n}=1)$ & 0.554 ± 0.12 & 0.919 ± 0.05 & 0.682 ± 0.09 &  0.770 ± 0.14 & 0.206 ± 0.05 & 0.321 ± 0.08 & 0.284 ± 0.07 & 0.864 ± 0.11 & 0.419 ± 0.07\\
 & & Partial $(\mathrm{n}=2)$ & 0.547 ± 0.11 & 0.920 ± 0.05 & 0.678 ± 0.08 & 0.751 ± 0.16 & 0.199 ± 0.06 & 0.312 ± 0.08 & 0.277 ± 0.08 & 0.861 ± 0.10 & 0.411 ± 0.08   \\
\hline
\hline
\multirow{2}{*}{Llama-2-13B Chat} & \multirow{2}{*}{0} & Exact $(\mathrm{n}=0)$ & 0.532 ± 0.17  & 0.968 ± 0.03  & 0.668 ± 0.15  & 0.432 ± 0.33  & 0.043 ± 0.05 & 0.077 ± 0.09 & 0.272 ± 0.12  & 0.994 ± 0.02  & 0.413 ± 0.14  \\
 & & Partial $(\mathrm{n}=1)$ & 0.562 ± 0.14 & 0.975 ± 0.02 & 0.701 ± 0.13 & 0.497 ± 0.27  & 0.047 ± 0.04 & 0.085 ± 0.07 & 0.279 ± 0.09 & 0.994 ± 0.01 & 0.428 ± 0.10   \\
 & & Partial $(\mathrm{n}=2)$ & 0.552 ± 0.13 & 0.980 ± 0.01 & 0.696 ± 0.12 & 0.598 ± 0.28 & 0.051 ± 0.05 & 0.092 ± 0.08 & 0.273 ± 0.06 & 0.990  ± 0.02 & 0.423 ± 0.08    \\  
\hline
\multirow{2}{*}{Llama-2-13B Chat} & \multirow{2}{*}{1} & Exact $(\mathrm{n}=0)$ &  0.517 ±  0.17 & 0.946 ±  0.05  & 0.646 ±  0.16  &  0.713 ±  0.31 & 0.082 ± 0.06  & 0.142 ±  0.10 & 0.251 ±  0.07  & 0.952 ±  0.11 & 0.392 ±  0.10  \\
 & & Partial $(\mathrm{n}=1)$ & 0.527 ±  0.15 & 0.958 ±  0.03 & 0.664 ±  0.14 & 0.686 ±  0.33  & 0.07 ±  0.05 & 0.121 ±  0.08 & 0.244 ±  0.03 & 0.985 ±  0.03 & 0.388 ±  0.05\\
 & & Partial $(\mathrm{n}=2)$ & 0.528 ±  0.15 & 0.963 ±  0.02 & 0.668 ±  0.13 & 0.733 ±  0.34 & 0.07 ±  0.04 & 0.124 ±  0.07 & 0.244 ±  0.03 & 0.985 ±  0.03 & 0.390 ±  0.04  \\
 \hline
\multirow{2}{*}{Llama-2-13B Chat} & \multirow{2}{*}{5} & Exact $(\mathrm{n}=0)$ & 0.537 ± 0.19   & 0.869 ± 0.12  & 0.648 ± 0.18  & 0.696 ± 0.33  & 0.096 ± 0.07  & 0.162 ± 0.12 & 0.254 ± 0.09  & \hlsb{0.997 ± 0.08} & 0.394 ± 0.11  \\
 & & Partial $(\mathrm{n}=1)$ & 0.556 ± 0.154 & 0.925 ± 0.08 & 0.680 ± 0.14 & 0.719 ± 0.32  & 0.084 ± 0.06 & 0.145 ± 0.10 & 0.260 ± 0.05 & \textbf{0.998 ± 0.01} & 0.409 ± 0.06 \\
 & & Partial $(\mathrm{n}=2)$ & 0.559 ± 0.156 & 0.931 ± 0.08 & 0.682 ± 0.14  & 0.716 ± 0.31 & 0.081 ± 0.06 & 0.140 ± 0.10 & 0.259 ± 0.05 & \hlc[aliceblue]{0.998 ± 0.06} & 0.408 ± 0.06  \\
\hline
\hline
\multirow{2}{*}{Llama-2-70B Chat} & \multirow{2}{*}{0} & Exact $(\mathrm{n}=0)$ & 0.513 ± 0.21  & 0.955 ± 0.11  & 0.647 ± 0.21  & 0.722 ± 0.27  & 0.074 ± 0.04  & 0.131 ± 0.07 & 0.313 ± 0.22  & 0.880 ± 0.22 & 0.428 ± 0.23  \\
 & & Partial $(\mathrm{n}=1)$ & 0.541 ± 0.17 & 0.997 ± 0.01 & 0.684 ± 0.15 & 0.722 ± 0.27  & 0.062 ± 0.04 & 0.113 ± 0.07 & \hlc[aliceblue]{0.341 ± 0.14} & 0.968 ± 0.07 & \hlsb{0.485 ± 0.15}\\
 & & Partial $(\mathrm{n}=2)$ & 0.530 ± 0.16 & 0.998 ± 0.01 &  0.678 ± 0.14 & 0.727 ± 0.28 & 0.069 ± 0.04 & 0.124 ± 0.07 & \textbf{0.355 ± 0.13} & 0.967 ± 0.08 & \textbf{0.504 ± 0.13}  \\
\hline
\multirow{2}{*}{Llama-2-70B Chat} & \multirow{2}{*}{1} & Exact $(\mathrm{n}=0)$ & 0.535 ± 0.15  & 0.939 ± 0.06  & 0.666 ± 0.13  & 0.876 ± 0.14  &  0.115 ± 0.04 & 0.198 ± 0.06 & 0.259 ± 0.13  & 0.931 ± 0.11 & 0.390 ± 0.15  \\
 & & Partial $(\mathrm{n}=1)$ & 0.544 ± 0.14  & 0.953 ± 0.05 & 0.681 ± 0.11 & 0.898 ± 0.11  & 0.120 ± 0.03  & 0.208 ± 0.05 & 0.264 ± 0.07 & 0.982 ± 0.03 &  0.410 ± 0.08\\
 & & Partial $(\mathrm{n}=2)$ & 0.528 ± 0.13 & 0.958 ± 0.04 & 0.670 ± 0.11 & 0.903 ± 0.10 & 0.118 ± 0.03 & 0.205 ± 0.05 & 0.267 ± 0.06 & 0.983 ± 0.03 & 0.415 ± 0.07  \\
\hline
\multirow{2}{*}{Llama-2-70B Chat} & \multirow{2}{*}{5} & Exact $(\mathrm{n}=0)$ & 0.578 ± 0.15  & 0.965 ± 0.03  & 0.709 ± 0.12  & 0.911 ± 0.13  & 0.177 ± 0.06  & 0.291 ± 0.10 & 0.253 ± 0.11  & 0.970 ± 0.08 & 0.388 ± 0.14  \\
 & & Partial $(\mathrm{n}=1)$ & 0.584 ± 0.12 & 0.965 ± 0.03 & 0.719 ± 0.10 & 0.923 ± 0.09  & 0.170 ± 0.05 & 0.283 ± 0.07 & 0.257 ± 0.04 & 0.976 ± 0.06 & 0.405 ± 0.05\\
 & & Partial $(\mathrm{n}=2)$ & 0.575 ± 0.12 & 0.961 ± 0.04 & 0.711 ± 0.09 & 0.917 ± 0.09 & 0.158 ± 0.05 & 0.265 ± 0.07 & 0.267 ± 0.04 & 0.985 ± 0.04 & 0.419 ± 0.04  \\
\hline
\hline
\end{tabular}
\label{tbl:few_shot_sentiment_prediction}
  \end{adjustbox}
\end{table*}

\end{document}